\documentclass[letterpaper,table]{article} 
\usepackage{aaai2026}
\usepackage{times}  
\usepackage{helvet}  
\usepackage{courier}  
\usepackage[hyphens]{url}  
\usepackage{graphicx} 
\usepackage{natbib}  
\usepackage{caption} 
\usepackage{algorithm}
\usepackage{algorithmic}
\usepackage{amsmath}  
\usepackage{textcomp}
\usepackage{booktabs,threeparttable,multirow,tabularx}
\usepackage{siunitx}
\usepackage{makecell}
\usepackage{rotating} 
\newcolumntype{d}{S[table-format=3.0]} 
\usepackage{placeins}

\usepackage{amssymb}

\usepackage{booktabs}
\newcolumntype{L}[1]{>{\raggedright\arraybackslash}p{#1}}
\newcolumntype{C}[1]{>{\centering\arraybackslash}p{#1}}
\newcolumntype{R}[1]{>{\raggedleft\arraybackslash}p{#1}}
\usepackage{array}
\newcolumntype{M}[1]{>{\centering\arraybackslash}m{#1}}

\usepackage{newfloat}
\usepackage{listings}
\DeclareCaptionStyle{ruled}{labelfont=normalfont,labelsep=colon,strut=off} 
\floatstyle{ruled}
\newfloat{listing}{tb}{lst}{}
\floatname{listing}{Listing}
\title{MCA-Bench: A Multimodal Benchmark for Evaluating CAPTCHA Robustness Against VLM-based Attacks}

\author{%
  Zonglin~Wu\textsuperscript{1}\quad
  Yule~Xue\textsuperscript{1}\quad
  Yaoyao~Feng\textsuperscript{1}\quad
  Xiaolong~Wang\textsuperscript{1}\quad
  Yiren~Song\textsuperscript{2}\textsuperscript{*}\\[3pt]
}
\affiliations{%
  \textsuperscript{1} Southwest University \\
  \textsuperscript{2} National University of Singapore\\
}

\usepackage{bibentry}

\begin{document}

\maketitle

\begin{abstract}
As automated attack techniques rapidly advance, CAPTCHAs remain a critical defense mechanism against malicious bots. However, existing CAPTCHA schemes encompass a diverse range of modalities—from static distorted text and obfuscated images to interactive clicks, sliding puzzles, and logic-based questions—yet the community still lacks a unified, large-scale, multimodal benchmark to rigorously evaluate their security robustness. To address this gap, we introduce MCA-Bench, a comprehensive and reproducible benchmarking suite that integrates heterogeneous CAPTCHA types into a single evaluation protocol. Leveraging a shared vision–language model backbone, we fine-tune specialized cracking agents for each CAPTCHA category, enabling consistent, cross-modal assessments. Extensive experiments reveal that MCA-Bench effectively maps the vulnerability spectrum of modern CAPTCHA designs under varied attack settings, and—crucially—offers the first quantitative analysis of how challenge complexity, interaction depth, and model solvability interrelate. Based on these findings, we propose three actionable design principles and identify key open challenges, laying the groundwork for systematic CAPTCHA hardening, fair benchmarking, and broader community collaboration. 
\begin{links}
    \link{Code}{https://github.com/noheadwuzonglin/MCA-Bench}
    \link{Datasets}{https://www.kaggle.com/datasets/luffy798/mca-benchmultimodal-captchas}
\end{links}
\end{abstract}
\begin{figure*}[t]
\centering
\includegraphics[width=0.82\textwidth]{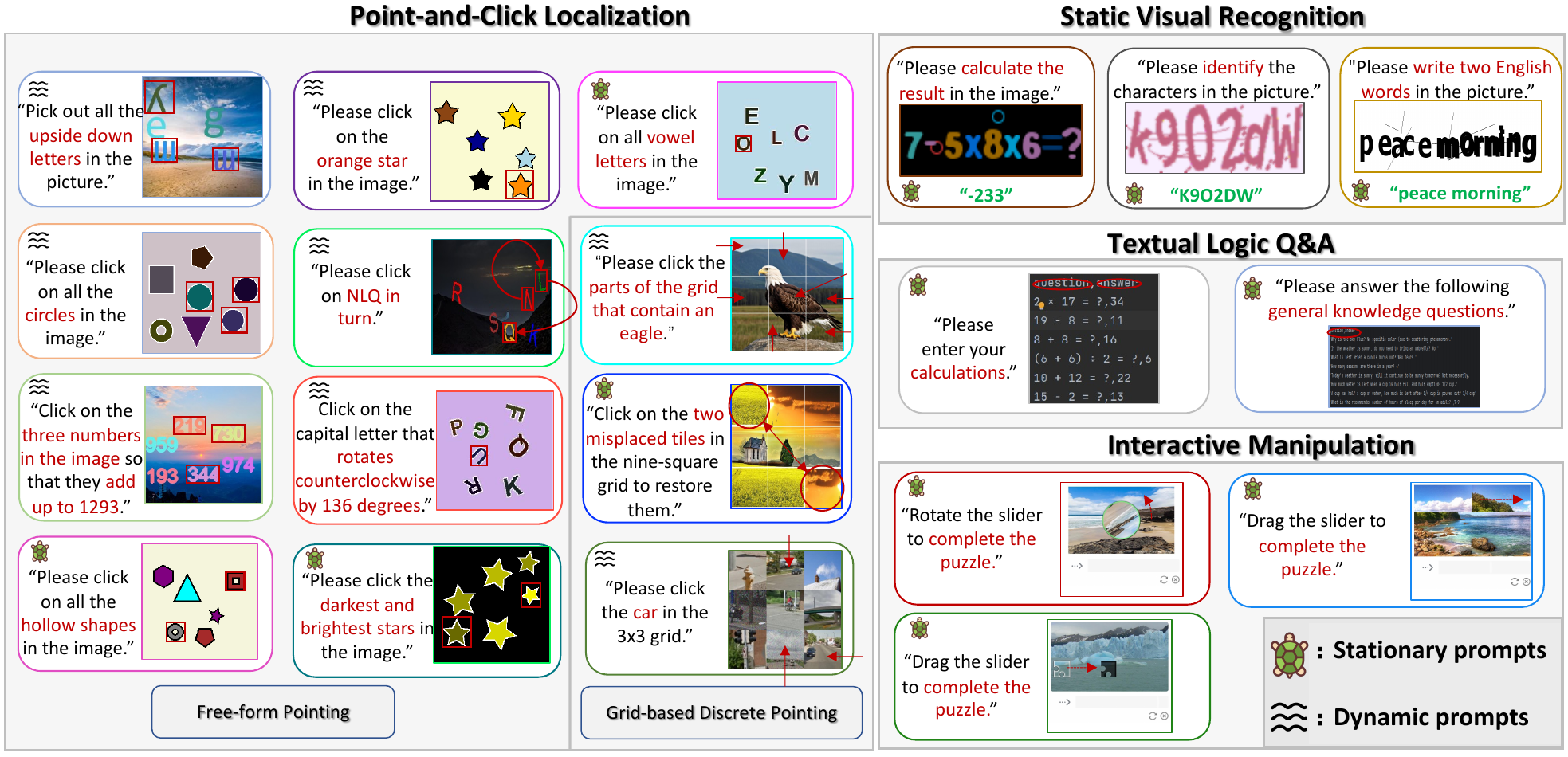} 
\caption{\textbf{Data samples from MCA-Bench.} Includes four categories and 20 sub-clusters of Point-and-Click Localization, Static Visual Recognition, Textual Logic Q\&A and Interactive Manipulation.}
\label{fig1.1}
\end{figure*}


\section{Introduction}
CAPTCHA (Completely Automated Public Turing test to tell Computers and Humans Apart) has been crucial in protecting online services from automated attacks. However, with advancements in deep learning, computer vision, and multimodal pretraining models, many CAPTCHA types once considered secure are now vulnerable to machine-learning-based attacks \cite{bursztein2011text,10.1007/3-540-58095-6_9,wang2023voyager}. Techniques such as GANs, vision-language models (VLMs), and reinforcement learning have enabled attackers to mimic human behavior with increasing precision \cite{10.5555/2974989,noury2020deep,kumar2021benchmarks,schick2023toolformer}. As a result, researchers have begun developing multimodal CAPTCHA datasets and evaluation frameworks to assess model performance across various CAPTCHA types \cite{acien2020becaptcha,farebrother2019generalization,acien2021becaptcha}. This makes it vital to reassess CAPTCHA’s real-world security to ensure a trustworthy Internet service \cite{10.1145/360018.360022,farebrother2019generalization}.

Existing studies often target specific CAPTCHA types without broad comparisons \cite{bursztein2011text,ci2024ringid,10.5555/3327345.3327436}. The absence of a large-scale, multimodal benchmark \cite{hernandez2021basecass,sanh2021multitask,ci2024wmadapter} limits systematic evaluation and hinders robust CAPTCHA design. A unified evaluation platform is urgently needed.

MCA-Bench is the first end-to-end CAPTCHA security benchmark spanning four modalities—static visual recognition, point-and-click localization, interactive manipulation and textual logic Q\&A—across twenty real-world tasks. It provides over 180000 training samples and a 4000-item test set, organized into four clusters that respectively evaluate OCR robustness to visual noise, target retrieval in complex scenes, human-like interaction behaviors, and multi-step language reasoning. Representative samples from the MCA-Bench dataset are shown in Figure \ref{fig1.1}.

We use Qwen2.5-VL-7B as the vision-language backbone, fine-tuned with LoRA adapters for each task \cite{hu2022lora}. Training for static and logic CAPTCHAs is supervised with target labels, while for interactive tasks, human demonstration data is used for behavior cloning. A specially designed JSON protocol facilitates large-scale evaluation and integration.

We use pass rate as the core metric. Evaluation shows multimodal VLMs exceed 96\% accuracy on simple tasks but fall to as low as 2.5\% on complex ones requiring physical interaction or multi-step reasoning. This reveals that visual confusion, interaction depth, and semantic complexity jointly drive attack difficulty, offering practical guidance for CAPTCHA design. MCA-Bench is open-sourced to enable reproduction and sustain iterative attack/defense benchmarking. The main contributions are as follows:
\begin{itemize}
\item MCA-Bench: the first large-scale, cross-modal CAPTCHA attack benchmark with 20 real tasks.
\item Proposed a unified evaluation pipeline with a single pass-rate metric and open source scripts.
\item First full-scale CAPTCHA security assessment with guidance for human–machine verification.
\end{itemize}

\section{Related Work}
\subsection{VLMs in Structured Image Tasks}
Visual Language Models (VLMs) unify image and text representations, with early models like CLIP \cite{noury2020deep} and ALIGN \cite{hossen2022aaecaptcha} performing well on simple tasks but lacking complex reasoning. Recent approaches integrate LLMs with visual encoders to enhance cross-modal understanding—e.g., BLIP-2 \cite{kumar2021benchmarks} employs frozen components for efficiency, while MiniGPT-4 \cite{Wang_2024} maps visual features to LLMs for improved QA. As focus shifts to structured inputs like tables and documents—similar to text-heavy, noisy CAPTCHA images—models such as Qwen-VL \cite{song2025idprotector} advance encoder design and multilingual layout understanding. Architectures like LLaVA \cite{li2023blip} and MiniGPT-4 leverage unified visual projection and instruction tuning to handle distortions and occlusions. However, most VLMs remain unadapted to CAPTCHA tasks, limiting their performance and generalization.
\subsection{Evolution of Intelligent Agents Toward AGI}
An AI agent is an autonomous entity that perceives, decides, and acts in its environment \cite{mckinney2020training}, characterized by autonomy, reactivity, and social interactivity—traits crucial to AGI development \cite{searles2023empirical,hong2024metagptmetaprogrammingmultiagent}. Agent research has evolved from symbolic rule-based systems \cite{luo2024image,10.5555/3305381.3305498}, limited by uncertainty and scalability \cite{radford2021learning}, to reactive agents with fast perception-action loops \cite{10.5555/110787} but limited planning. Reinforcement learning enabled agents like AlphaGo \cite{shah2023lm}, though sample inefficiency and poor generalization persisted \cite{park2023generative,pmlr-v139-jia21b}. Transfer and meta-learning improved adaptation via knowledge reuse \cite{10.5555/2772879.2772905,nian2022deep,deng2024oedipus,Fakoor2020}, despite high pre-training costs \cite{elson2007asirra}. Recently, LLM-based agents show emergent reasoning, planning \cite{liu2024image,achiam2023gpt,anthony_therrien_2024}, multimodal understanding (e.g., BLIP-2 \cite{1996cs........5103K}), dynamic task decomposition (e.g., Voyager \cite{sivakorn2016robot}), and tool use (e.g., Toolformer \cite{10.1023/A:1015008417172}), enabling general-purpose intelligence \cite{russell2021modern,song2024anti}. Their zero-shot generalization \cite{sumers2024cognitivearchitectureslanguageagents} and social collaboration abilities \cite{1571980075547310080} mark a paradigm shift in agent research.

\subsection{Advances and Challenges in CAPTCHA Security}
Early text CAPTCHAs relied on heavy distortion and noise, but CNN-based segmentation attacks soon defeated many schemes, including reCAPTCHA \cite{chellapilla2005building,shet2014you,bursztein2011text}. Image-based CAPTCHAs, such as ASIRRA, emphasized visual cognition yet were ultimately bypassed by SVM classifiers trained on public datasets \cite{ding2025illusioncaptcha,gao2021research}. Modern defenses now employ deep object detection, diffusion models, and style transfer for increased complexity \cite{mann2020language,van2023anti,109c6f85f2734b649d5f7b7a1946b2c1,ci2024ringid,10.5555/2974989,NEURIPS2023_6dcf277e}. Interactive CAPTCHAs (e.g., reCAPTCHA v2) add user gestures like clicking or dragging to resist automation \cite{10.5555/1625995.1626091}. Nonetheless, challenges remain in adaptive difficulty, device consistency, adversarial robustness, and reproducible benchmarks \cite{jiang2023diff,alsuhibany2016benchmark}.

\FloatBarrier   
\begin{table*}[t]
\centering

\fontsize{9pt}{11pt}\selectfont
\setlength{\tabcolsep}{3pt}

\begin{tabular*}{\textwidth}{@{\extracolsep{\fill}}cccccccccccc}
\hline
Model & Pass@k &
\makecell{3×3\\grid sel.} & \makecell{3×3\\jig-swap} & \makecell{Arith.\\char.} & \makecell{Arith.\\sel.} &
\makecell{Hollow\\pattern} & \makecell{Distort.\\word} & \makecell{Classic\\char.} & \makecell{Sequential\\letter} &
\makecell{Bright.\\dist.} & \makecell{Sliding\\block} \\
\hline
\multirow{4}{*}{ChatGPT-4o} 
 & Pass@2 & 0.780 & 0.010 & 0.360 & 0.015 & 0.980 & 0.920 & 0.045 & 0.840 & 0.520 & 0.220 \\
 & Pass@3 & 0.780 & 0.010 & 0.360 & 0.020 & 0.985 & 0.920 & 0.050 & 0.840 & 0.520 & 0.225 \\
 & Pass@4 & 0.780 & 0.010 & 0.360 & 0.020 & 0.985 & 0.920 & 0.050 & 0.840 & 0.520 & 0.225 \\
 & Pass@5 & 0.780 & 0.015 & 0.360 & 0.020 & 0.990 & 0.920 & 0.050 & 0.840 & 0.520 & 0.225 \\
\hline
\multirow{4}{*}{Seed1.5-VL} 
 & Pass@2 & 0.800 & 0.005 & 0.320 & 0.020 & 0.865 & 0.960 & 0.085 & 0.820 & 0.465 & 0.185 \\
 & Pass@3 & 0.800 & 0.005 & 0.320 & 0.020 & 0.865 & 0.960 & 0.085 & 0.825 & 0.465 & 0.190 \\
 & Pass@4 & 0.805 & 0.005 & 0.320 & 0.020 & 0.870 & 0.960 & 0.085 & 0.825 & 0.465 & 0.195 \\
 & Pass@5 & 0.805 & 0.005 & 0.320 & 0.020 & 0.875 & 0.960 & 0.085 & 0.825 & 0.465 & 0.195 \\
\hline
\multirow{4}{*}{Gemini2.5-Pro} 
 & Pass@2 & 0.820 & 0.005 & 0.360 & 0.015 & 0.940 & 0.920 & 0.140 & 0.905 & 0.480 & 0.230 \\
 & Pass@3 & 0.825 & 0.005 & 0.360 & 0.015 & 0.945 & 0.920 & 0.140 & 0.905 & 0.480 & 0.235 \\
 & Pass@4 & 0.825 & 0.010 & 0.360 & 0.020 & 0.945 & 0.920 & 0.145 & 0.910 & 0.485 & 0.235 \\
 & Pass@5 & 0.825 & 0.010 & 0.360 & 0.020 & 0.945 & 0.920 & 0.150 & 0.910 & 0.485 & 0.235 \\
\hline
\multirow{4}{*}{Original Qwen-2.5vl-7B-Instr.} 
 & Pass@2 & 0.660 & 0.010 & 0.365 & 0.010 & 0.940 & 0.020 & 0.025 & 0.340 & 0.220 & 0.100 \\
 & Pass@3 & 0.660 & 0.010 & 0.365 & 0.010 & 0.940 & 0.020 & 0.025 & 0.340 & 0.220 & 0.100 \\
 & Pass@4 & 0.665 & 0.015 & 0.365 & 0.010 & 0.950 & 0.025 & 0.030 & 0.345 & 0.225 & 0.100 \\
 & Pass@5 & 0.665 & 0.015 & 0.365 & 0.010 & 0.950 & 0.025 & 0.030 & 0.345 & 0.225 & 0.100 \\
\hline
\multirow{4}{*}{Human} 
 & Pass@2 & 0.960 & 0.980 & 0.990 & 0.955 & 0.995 & 0.885 & 0.930 & 0.960 & 0.965 & 0.980 \\
 & Pass@3 & 0.985 & 0.985 & 0.990 & 0.955 & 0.995 & 0.965 & 0.950 & 0.980 & 0.985 & 0.985 \\
 & Pass@4 & 0.995 & 0.985 & 0.995 & 0.965 & 1.000 & 0.985 & 0.980 & 0.985 & 0.985 & 0.990 \\
 & Pass@5 & 1.000 & 0.990 & 1.000 & 0.970 & 1.000 & 0.990 & 0.995 & 0.995 & 0.990 & 0.995 \\
\hline
\multirow{4}{*}{Finetuned Qwen-2.5vl-7B-Instr.} 
 & Pass@2 & 0.965 & 0.015 & 0.700 & 0.025 & 0.995 & 0.985 & 0.325 & 0.925 & 0.665 & 0.365 \\
 & Pass@3 & 0.970 & 0.015 & 0.705 & 0.025 & 0.995 & 0.985 & 0.325 & 0.925 & 0.665 & 0.365 \\
 & Pass@4 & 0.975 & 0.015 & 0.705 & 0.030 & 0.995 & 0.985 & 0.325 & 0.925 & 0.670 & 0.365 \\
 & Pass@5 & 0.980 & 0.020 & 0.710 & 0.035 & 1.000 & 0.995 & 0.330 & 0.930 & 0.675 & 0.365 \\
\hline
Model & Pass@k &
\makecell{Align.\\sliders} & \makecell{Rotate\\block} & \makecell{Geom.\\shape} & \makecell{Rotat.\\letter} &
\makecell{Color\\discr.} & \makecell{Vowel\\sel.} & \makecell{Full-img\\grid sel.} & \makecell{Text-based\\arith.} &
\makecell{Common\\sense} & \makecell{Invert.\\letter} \\
\hline
\multirow{4}{*}{ChatGPT-4o} 
 & Pass@2 & 0.420 & 0.140 & 0.440 & 0.100 & 0.960 & 0.540 & 0.240 & 0.980 & 0.925 & 0.425 \\
 & Pass@3 & 0.420 & 0.140 & 0.440 & 0.100 & 0.960 & 0.540 & 0.240 & 0.985 & 0.925 & 0.425 \\
 & Pass@4 & 0.420 & 0.140 & 0.440 & 0.100 & 0.960 & 0.540 & 0.245 & 0.985 & 0.925 & 0.430 \\
 & Pass@5 & 0.420 & 0.140 & 0.440 & 0.100 & 0.960 & 0.540 & 0.245 & 0.985 & 0.930 & 0.430 \\
\hline
\multirow{4}{*}{Seed1.5-VL} 
 & Pass@2 & 0.380 & 0.180 & 0.840 & 0.140 & 0.930 & 0.880 & 0.230 & 0.905 & 0.840 & 0.360 \\
 & Pass@3 & 0.380 & 0.180 & 0.840 & 0.140 & 0.930 & 0.880 & 0.235 & 0.910 & 0.845 & 0.365 \\
 & Pass@4 & 0.380 & 0.185 & 0.840 & 0.140 & 0.930 & 0.880 & 0.235 & 0.920 & 0.845 & 0.365 \\
 & Pass@5 & 0.380 & 0.185 & 0.845 & 0.140 & 0.935 & 0.880 & 0.235 & 0.920 & 0.850 & 0.365 \\
\hline
\multirow{4}{*}{Gemini2.5-Pro} 
 & Pass@2 & 0.420 & 0.160 & 0.845 & 0.120 & 0.925 & 0.900 & 0.280 & 0.880 & 0.860 & 0.385 \\
 & Pass@3 & 0.420 & 0.160 & 0.845 & 0.120 & 0.925 & 0.900 & 0.280 & 0.885 & 0.865 & 0.385 \\
 & Pass@4 & 0.420 & 0.165 & 0.845 & 0.120 & 0.925 & 0.905 & 0.285 & 0.890 & 0.870 & 0.390 \\
 & Pass@5 & 0.420 & 0.165 & 0.850 & 0.125 & 0.930 & 0.905 & 0.290 & 0.895 & 0.875 & 0.400 \\
\hline
\multirow{4}{*}{Original Qwen-2.5vl-7B-Instr.} 
 & Pass@2 & 0.300 & 0.060 & 0.885 & 0.100 & 0.700 & 0.760 & 0.140 & 0.960 & 0.920 & 0.365 \\
 & Pass@3 & 0.305 & 0.060 & 0.885 & 0.100 & 0.700 & 0.760 & 0.140 & 0.960 & 0.920 & 0.365 \\
 & Pass@4 & 0.305 & 0.060 & 0.885 & 0.100 & 0.700 & 0.760 & 0.140 & 0.965 & 0.925 & 0.365 \\
 & Pass@5 & 0.305 & 0.065 & 0.885 & 0.100 & 0.700 & 0.760 & 0.145 & 0.965 & 0.925 & 0.370 \\
\hline
\multirow{4}{*}{Human} 
 & Pass@2 & 0.975 & 0.965 & 0.990 & 0.875 & 0.970 & 0.960 & 0.940 & 0.970 & 0.875 & 0.955 \\
 & Pass@3 & 0.995 & 0.990 & 0.990 & 0.950 & 0.975 & 0.980 & 0.975 & 0.985 & 0.905 & 0.985 \\
 & Pass@4 & 1.000 & 0.995 & 0.995 & 0.985 & 0.990 & 0.995 & 0.990 & 0.990 & 0.955 & 1.000 \\
 & Pass@5 & 1.000 & 0.995 & 1.000 & 0.990 & 0.995 & 1.000 & 0.995 & 1.000 & 0.985 & 1.000 \\
\hline
\multirow{4}{*}{Finetuned Qwen-2.5vl-7B-Instr.} 
 & Pass@2 & 0.565 & 0.285 & 0.960 & 0.335 & 0.990 & 0.975 & 0.360 & 0.985 & 0.975 & 0.520 \\
 & Pass@3 & 0.565 & 0.285 & 0.960 & 0.335 & 0.990 & 0.975 & 0.360 & 0.985 & 0.975 & 0.520 \\
 & Pass@4 & 0.565 & 0.285 & 0.960 & 0.335 & 0.995 & 0.975 & 0.370 & 0.990 & 0.980 & 0.525 \\
 & Pass@5 & 0.570 & 0.285 & 0.960 & 0.340 & 0.995 & 0.980 & 0.370 & 0.995 & 0.990 & 0.530 \\
\hline
\end{tabular*}
\caption{\textbf{Performance of vision-language models on CAPTCHA-Bench.} This table presents Pass@k accuracy on 20 CAPTCHA task types, comparing models and human performance.}
\label{tab:captcha_bench}
\end{table*}
\FloatBarrier

\begin{figure*}[t]
\centering
\includegraphics[width=0.82\textwidth]{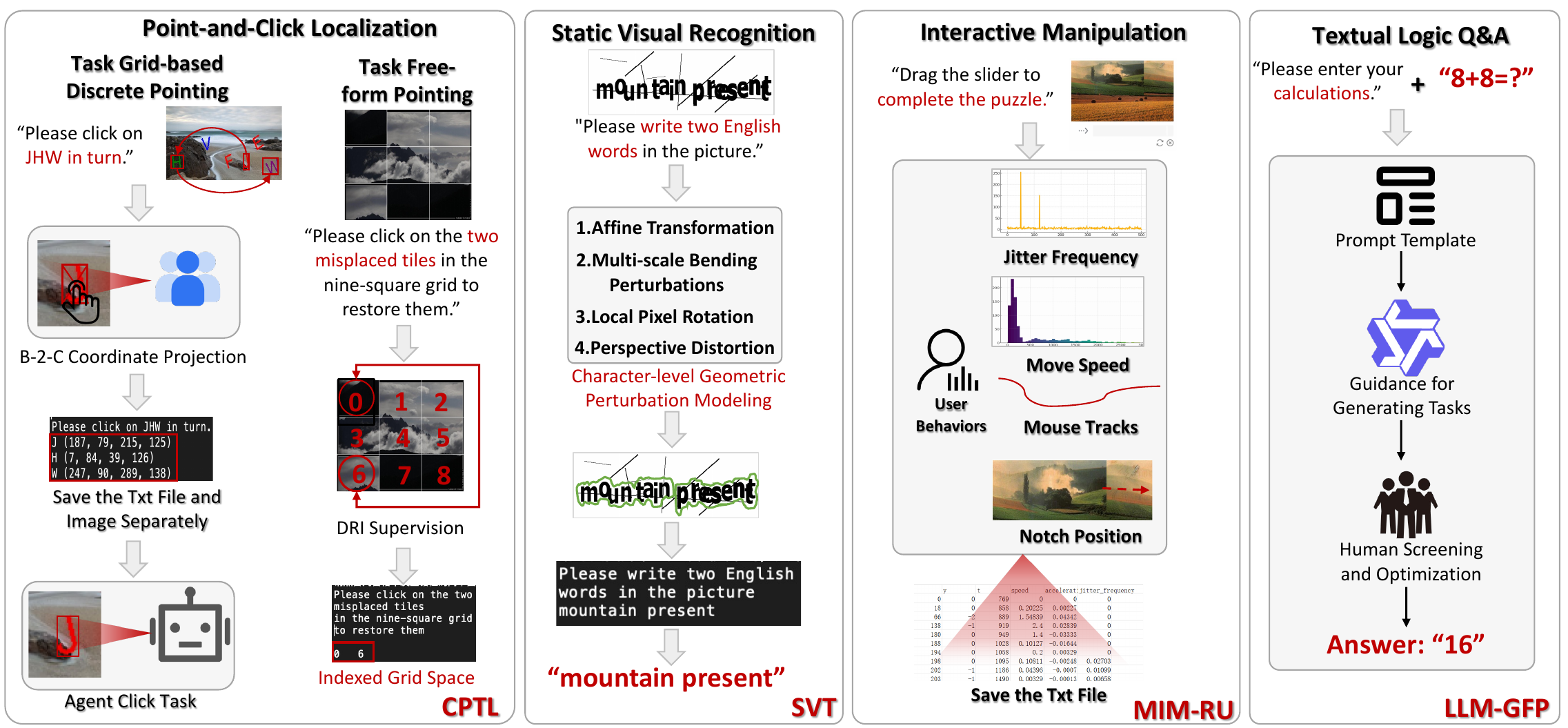} 
\caption{\textbf{Schematic overview of the MCA-Bench data-acquisition and annotation workflow.} From left to right, the four grey panels are Static Visual Recognition, Interactive Manipulation, Point-and-Click Localization, and Textual Logic Q\&A; the red labels mark each category’s data-collection pipeline. Each pipeline has four stages: (i) define the raw input format; (ii) apply task-specific geometric transforms, coordinate projections, or prompt/template generation; (iii) separate fine-grained annotation types; and (iv) save the annotations to text files.}
\label{fig1.2}
\end{figure*}

\begin{figure*}[tbp]
\centering
\includegraphics[width=0.82\textwidth]{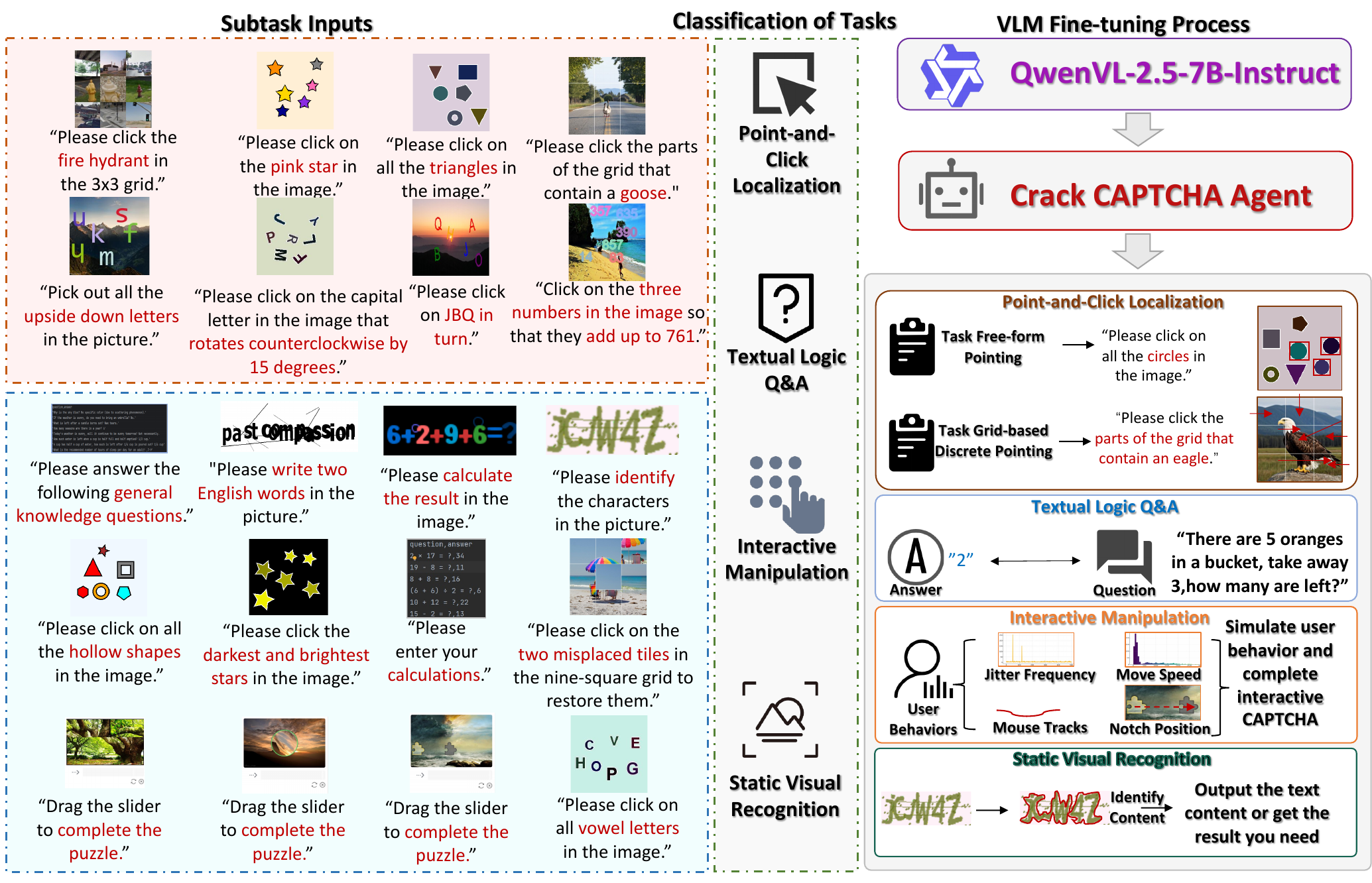} 
\caption{\textbf{Schematic of Data Flow Across Four Framework Stages.} The schematic diagram illustrates the data flow and key module configuration across the four stages of the end-to-end framework: unified interface access, gent fine-tuning loading, collaborative inference execution, structured result feedback.}
\label{fig2}
\end{figure*}

\begin{figure*}[t]
\centering
\includegraphics[width=0.9\textwidth]{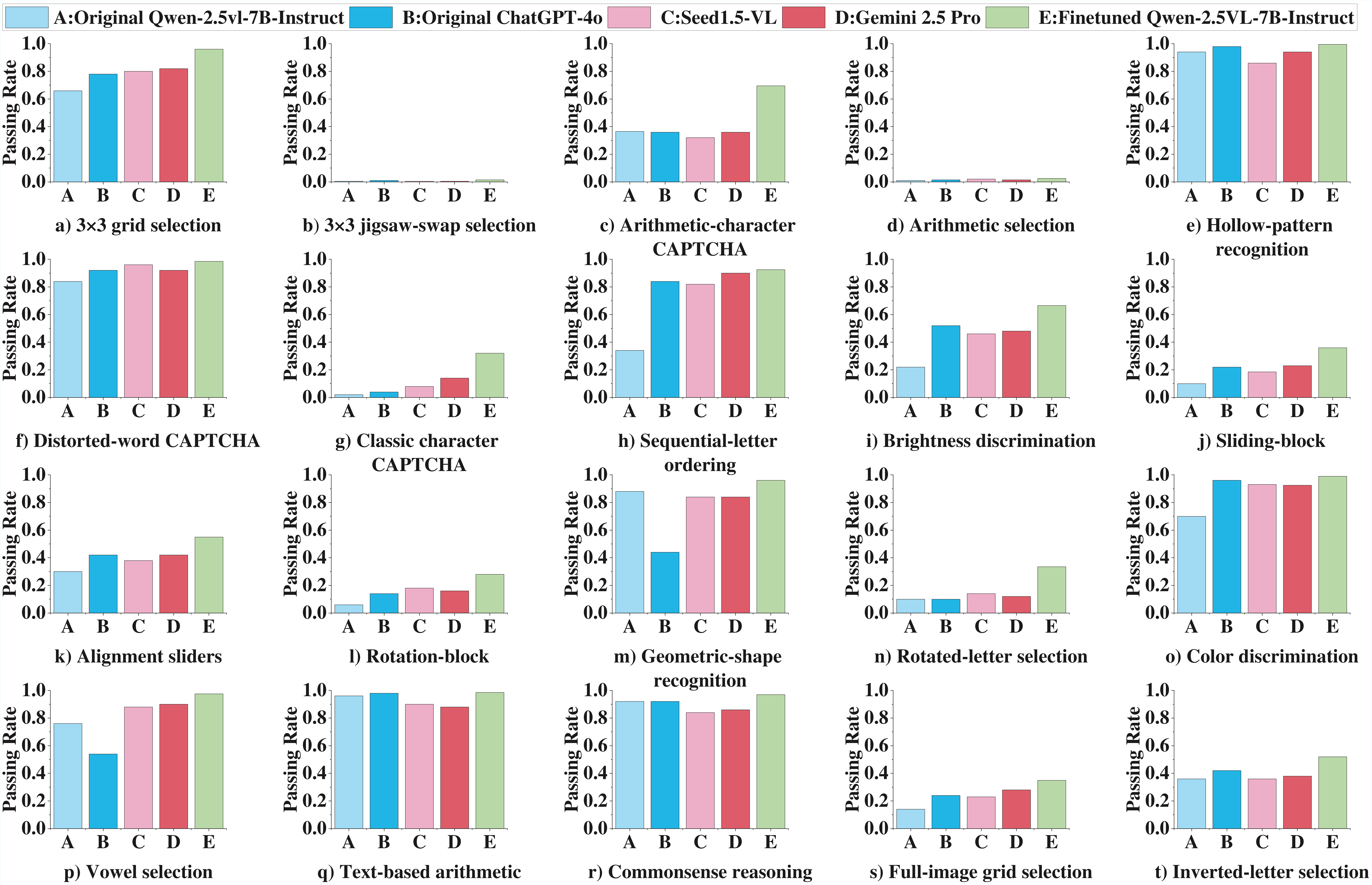} 
\caption{\textbf{Performance Comparison of Multimodal Language Models on MCA-Bench CAPTCHA Tasks.} The figure compares the success rates of models including Qwen2.5-VL-7B-Instruct, ChatGPT-4o, Seed1.5-VL, Gemini2.5-Pro, and fine-tuned Qwen2.5-VL-7B-Instruct across MCA-Bench CAPTCHA tasks, covering basic visual recognition, character-based recognition, and advanced multi-step reasoning challenges. Results show fine-tuning consistently improves performance, yet even top-performing models lag behind human-level robustness in complex reasoning tasks.}
\label{fig4}
\end{figure*}

\section{Dataset Construction Process}
This section introduces the MCA Bench pipeline—from raw sample collection to final release—detailing processing and annotation strategies across four task clusters and twenty subtasks. See Figure~\ref{fig1.2} for the complete workflow.
\subsection{Data Collection Sources and Compliance Strategies}
In developing the MCA-Bench benchmark suite, we implemented a comprehensive data collection and compliance management system to uphold scientific integrity and ensure legal and ethical standards in evaluating CAPTCHA security. Our approach combines independently created data and reused public datasets. For the former, we designed a diverse CAPTCHA dataset, including distorted text, obfuscated graphics, interactive tasks, and puzzles, based on a thorough analysis of technology trends and security needs. These designs are original, validated, and proprietary to our team.
For reused datasets, we carefully selected authorized academic, open-source, and industry resources, adhering to licensing agreements and privacy regulations. We ensured compliance through preprocessing steps like anonymization and mitigating privacy risks. Additionally, we established a multi-level review system with legal and ethics oversight, encrypted storage, and tiered access control to guarantee ongoing compliance and data security for MCA-Bench.
\subsection{Data Collection and Processing}
\subsubsection{Text-based Task Data Collection}
For text-based CAPTCHA tasks, we designed a semi-automated, LLM-driven pipeline to efficiently generate and filter math and commonsense questions, minimizing manual effort while preserving quality and diversity. Using adaptive prompts, the Qwen LLM produced structured, semantically relevant QA pairs, guided by knowledge constraints, task-aware sampling, and type-controlled difficulty. Outputs were refined via manual filtering with a custom evaluation protocol assessing grammar, reasoning, and ambiguity to ensure clarity and robustness for real-world deployment.
\subsubsection{Click-based Coordinate Task Data Collection}
We propose CPTL (Click-based Positioning and Target Localization), a multimodal benchmark designed to evaluate models’ spatial localization and image-language alignment across varying complexities. It consists of two tasks: Free-form Pointing and Grid-based Discrete Pointing. In Free-form Pointing, we combine procedural background perturbations with public datasets, such as Flickr scenic photos, to create images rich in semantic content. The Grid-based Discrete Pointing task uses a 3×3 grid to evaluate decision-making within specific regions, dividing the image into 9 segments (0–8). The image content includes a 64-class animal dataset, Flickr scenic backgrounds, Google ReCaptcha V2 challenges, and manually designed targets.

CPTL enables adjustable evaluation of spatial complexity and instruction modalities, effectively isolating spatial parsing from language comprehension. It simulates real-world CAPTCHA conditions, testing model robustness against noisy backgrounds, multi-target interference, and weak prompts. Experiments demonstrate that models fine-tuned with CPTL achieve strong localization and semantic understanding, establishing a reliable benchmark for future CAPTCHA security research.
\subsubsection{Static Visual Recognition Task Data Collection}
We propose SVT (Static Visual Textual Understanding), a benchmark to evaluate multimodal models’ ability to recognize and reconstruct text from distorted images. SVT leverages procedural image generation and linguistic constraints, applying character-level geometric distortions, stochastic noise, and occlusions to create challenging yet interpretable samples. It assesses models’ fine-grained visual attention and robustness to structural noise, spanning from perception to symbolic reasoning. Experiments demonstrate SVT’s effectiveness in revealing limitations of pretrained multimodal models in character-level understanding.
\subsubsection{Interactive Behavior Task Data Collection}
We propose a multimodal interaction modeling framework for interactive CAPTCHA tasks, leveraging real user data to enhance vision-language models' dynamic understanding. Focusing on continuous motion interactions—such as sliding alignment, rotation calibration, and trajectory restoration—the framework requires models to interpret spatial structures, motion directions, and behavioral patterns. We collected diverse user trajectories, annotated with timestamps and velocity, enabling realistic spatiotemporal modeling.

Tasks are structured around target-state restoration, divided into standardized subtasks with clear start, goal, and intermediate states. Interaction trajectories are serialized as behavioral vectors combined with visual data for joint learning, significantly enhancing action alignment. To evaluate models comprehensively, we propose new metrics: Center Deviation Error, Angular Restoration Accuracy, Slide Path Alignment Rate, and Motion Variability Index.
\subsection{Data Annotation Strategy}
We built a standardized, task-driven annotation framework with four representative task types to ensure consistent multimodal CAPTCHA training and evaluation.
\subsubsection{Unified Intent Modeling for Coordinate Pointing and Grid Selection}
For free-form coordinate pointing tasks, we adopt a box-to-center projection strategy. Annotators label each target by marking the top-left and bottom-right corners of its bounding box using absolute pixel coordinates, with $(0,0)$ at the image’s top-left. The geometric center of the box serves as the training target. During inference, a prediction is considered correct if it falls within the box, following an IoB-Gated Validation rule. This provides spatial tolerance, improves robustness to outliers, and stabilizes training. Formally, the validation criterion for a predicted point $p$ and bounding box $b = [\,b_{\min}, b_{\max}\,]$ is defined as:
\begin{equation}
   \mathcal{G}\bigl(p,\,b\bigr) \;=\; \mathbb{I}\Bigl(\bigl\lVert\,D^{-1}\bigl(p - \tfrac{1}{2}(b_{\min} + b_{\max})\bigr)\bigr\rVert_{\infty} \leq \tfrac{1}{2}\Bigr)
\label{eq:iob-val}
\end{equation}

Here, $\mathbb{I}(\cdot)$ is the indicator function (1 if the condition holds, 0 otherwise),
$D = \operatorname{diag}(\,{b}_{\max} - \,{b}_{\min})$ is the diagonal matrix of bounding‐box width and height,
and $\lVert \cdot \rVert_{\infty}$ is the $\ell_\infty$ norm (the maximum absolute component).
Thus, the expression tests whether the predicted point lies within the normalized box region centered on the ground‐truth box.

We propose a discrete region‐index supervision scheme: each image is split into a 3×3 row-major grid, and the target is classified by its cell index, aligning language to local regions and improving robustness under multi-object interference and ambiguous or similar visual cues.
\subsubsection{Static Visual Recognition Class Task Labeling Methods}
A unified labeling framework adapts to any static-vision task with task-specific supervision: for character recognition, we generate precise, content-aligned tags via automated text generation plus human review to balance speed and accuracy; for higher-level semantics, we use Visual-Symbolic Abstraction Parsing to deterministically translate visual structure into “final semantic solutions,” enabling rigorous logical reasoning and computation—yielding clear, consistent labels that drive precise model training and reliable evaluation even in the most complex cases.
\subsubsection{Interactive Behavior Task Labeling Methods}
We model interaction-based CAPTCHAs by capturing fine-grained user trajectories, target actions, and motion cues—whether sliding, rotating, or aligning layers. By fusing visual restoration with behavioral analysis, the system flags genuine human input (characterized by nonlinear paths, variable speed, and natural timing) and filters out the uniform, mechanically precise patterns typical of bots.

In slider CAPTCHAs, the user drags a puzzle piece to its gap; in rotation CAPTCHAs, they turn the image with a slider; in alignment CAPTCHAs, they slide layers until they coincide. Ten checkpoints record, and we score each trace for path smoothness, velocity profile, and plausible duration. For rotation, the slider’s horizontal travel $\Delta x\in[0,L]$ ($L=250$ px) maps to the counter-clockwise angle and can be computed as follows:
\begin{equation}
\phi(\Delta x) = \arg \left(\exp \left(i \cdot 2\pi \cdot \frac{\Delta x}{L} \right) \right) \cdot \frac{360^\circ}{2\pi}
\end{equation}

$\Delta x\in[0,L]$ is the slider’s horizontal displacement in pixels (with $L=250\,$px), $i^2=-1$ the imaginary unit; $\exp\bigl(i \cdot 2\pi \cdot \frac{\Delta x}{L})$ maps this displacement onto the complex unit circle, $\arg(\cdot)$ extracts its phase (in radians), and the factor $\frac{360^\circ}{2\pi}$ converts that phase into the rotation angle $\phi(\Delta x)$.

\section{Agent Pipeline and Experiments}

This section presents a detailed analysis of the experimental results based on the MCA-Bench benchmark.
\subsection{Set Up}

We fine‑tune LoRA adapters on Qwen2.5‑VL‑7B‑Instruct across 20 CAPTCHA tasks \cite{hu2022lora}. Inputs are 224×224 images paired with structured prompts. Training runs on 4 H20 GPUs (batch size 8, gradient accumulation 4 → effective 32) using AdamW with a linear LR decay from from $1\times10^{-4}$. We save checkpoints every 100 steps and stop early if validation loss doesn’t improve over 20 evaluations.
\begin{table}[H]
\centering

\small
\setlength{\tabcolsep}{4pt}
\begin{tabular}{@{}lcc@{}}
\toprule
\textbf{Dataset} & \textbf{Qwen2.5-VL-7B} & \textbf{Human} \\
\midrule

\multicolumn{3}{c}{\textbf{Point-and-click localization}} \\
\cmidrule{1-3}
3×3 grid selection           & 0.960 & 0.880 \\
Inverted-letter selection    & 0.520 & 0.940 \\
Geometric-shape recognition  & 0.960 & 0.980 \\
Brightness discrimination    & 0.665 & 0.780 \\
Hollow-pattern recognition   & 0.995 & 0.985 \\
Sequential-letter ordering   & 0.925 & 0.980 \\
Full-image grid selection    & 0.350 & 0.740 \\
Color discrimination         & 0.990 & 0.885 \\
Vowel selection              & 0.975 & 0.805 \\
Arithmetic selection         & 0.025 & 0.780 \\
Rotated-letter selection     & 0.335 & 0.745 \\
3×3 jigsaw-swap selection    & 0.015 & 0.805 \\

\midrule
\multicolumn{3}{c}{\textbf{Static visual recognition}} \\
\cmidrule{1-3}
Classic character CAPTCHA    & 0.320 & 0.920 \\
Distorted-word CAPTCHA       & 0.985 & 0.840 \\
Arithmetic-character CAPTCHA & 0.695 & 0.985 \\

\midrule
\multicolumn{3}{c}{\textbf{Interactive manipulation}} \\
\cmidrule{1-3}
Sliding-block                & 0.360 & 0.740 \\
Rotation-block               & 0.280 & 0.760 \\
Alignment sliders            & 0.550 & 0.720 \\

\midrule
\multicolumn{3}{c}{\textbf{Textual logic Q\&A}} \\
\cmidrule{1-3}
Text-based arithmetic        & 0.985 & 0.970 \\
Commonsense reasoning        & 0.970 & 0.860 \\

\bottomrule
\end{tabular}
\caption{\textbf{Comparison of Pass Rates for CAPTCHA Types:} Qwen2.5-VL-7B vs. Human Performance.}
\label{table:captcha_passrate}
\end{table}

\subsection{MCA-Bench: Cracking Capability Dataset}
Fine-tuning QWen-2.5VL-7B-instruct on MCA-Bench yields substantial gains. As shown in Fig \ref{fig4}, the adapted model surpasses its zero-shot baseline and closed-source peers across CAPTCHA tasks, enhancing visual recognition, logical reasoning, interaction, and robustness to complex challenges.
\subsection{MCA-Bench Dataset Overview}
MCA-Bench is the first multimodal CAPTCHA dataset covering visual recognition, point selection, textual reasoning, and interactive operations. It features varied image sizes, publicly sourced tasks for diversity and reproducibility, and a large training set that supports multi-image fine-tuning to improve generalization. 
\subsection{Agent System Pipeline}
The pipeline packs every image, prompt, and user action into one JSON record. A task-ID in the header activates the matching LoRA agent \cite{hu2022lora}, which adds its lightweight adapter to a shared frozen backbone. Visual- and text-embeddings flow through that agent to generate outputs—coordinates, strings, mouse traces, or character codes. These results are written back into the original JSON format for downstream use. This design lets the same backbone handle all CAPTCHA tasks with minimal memory and no intermediate parsers (Fig. \ref{fig2}).
\subsection{Fine-Tuned Model Performance in Visual Tasks}
On simple visual tasks with minimal noise, the fine-tuned VLM surpasses human speed and accuracy. However, its performance sharply declines under complex transformations such as distortion or rotation, highlighting VLMs' limitations compared to robust human perception.

\subsection{Zero-Shot vs. Fine-Tuned Performance Gaps}
Despite differing architectures and training regimes, zero-shot pass rates vary widely across Qwen2.5-VL, ChatGPT-4o, Seed1.5-VL, and Gemini2.5-Pro (Table \ref{tab:captcha_bench}), yet none match the fluid adaptability of human solvers. Fine-tuning on MCA-Bench consistently improves performance—especially for Qwen2.5-VL—but even the best-tuned agents remain below human robustness on multi-step reasoning or precise interaction tasks. This highlights that, while instruction tuning and backbone advances bring notable gains, human proficiency in nuanced, context-rich challenges is still unmatched.

\subsection{Human vs. AI in Reasoning and Interaction}
In tasks requiring reasoning or behavioral interaction—such as sliding puzzles, rotating tiles, multi-step reasoning, and common-sense judgment—the fine-tuned model still lags behind human performance. As shown in Table \ref{table:captcha_passrate}, despite improvements in standard visual recognition, the fine-tuned Qwen-Agent struggles with tasks demanding deeper understanding, contextual reasoning, and precise coordination.

\section{Conclusion}
MCA-Bench is the first unified benchmark showing that although LoRA-based attack agents achieve over 96\% accuracy on visual and shallow-semantic CAPTCHAs, their performance drops below 2.5\% on tasks involving physical interaction or multi-step reasoning—highlighting the limits of single-dimensional obfuscation. To address this gap, we introduce a tightly integrated human–machine verification paradigm built on three principles: (1) deep modality coupling — jointly fusing visual cue localization, logical inference, and interactive input in an adaptively evolving pipeline; (2) behavior-anchored validation — identifying humans through interaction trajectories such as timing, continuity, and subtle irregularities; and (3) session-specific semantic personalization — injecting unique semantic context into each challenge to prevent reuse or precomputation. Together, these principles create a cognitively rich, adaptive verification framework that substantially strengthens robustness against advanced AI-driven attacks.

\FloatBarrier
\bibliography{aaai2026}

\clearpage
\appendix
\onecolumn
\begin{center}
{\Large\bfseries MCA-Bench: A Multimodal Benchmark for Evaluating CAPTCHA Robustness Against VLM-based Attacks}

\vspace{0.5em}

{\Large Supplementary Material}

\vspace{3em}
{\Large\bfseries Outline}
\end{center}

\noindent\textcolor{black}{\rule{\textwidth}{0.5pt}}

\textbf{A\quad More Details on MCA-Bench} 

\hspace{3em}A.1\quad Overview of Data Sets 

\hspace{3em}A.2\quad Distribution Pattern and Training Balance Details for Each Type of Task 

\hspace{3em}A.3\quad Image Resolution Hierarchy 

\vspace{0.5em}
\textbf{B\quad Market Research on Mainstream CAPTCHA Datasets}

\hspace{3em}B.1\quad Comparative Analysis of User Experience and Security Performance 

\hspace{3em}B.2\quad Evaluation of Mainstream CAPTCHA Types

\hspace{3em}B.3\quad Application Trends and User Acceptance Analysis of Mainstream CAPTCHA Types 

\hspace{3em}B.4\quad Evaluation and Analysis of Mainstream CAPTCHA Types Selection

\hspace{3em}B.5\quad Analysis of CAPTCHA Error-Proneness and User Self-Recovery

\hspace{3em}B.6\quad Semiotic Analysis of CAPTCHA Types

\vspace{0.5em}
\textbf{C\quad More Extensive Experiments}

\hspace{3em}C.1\quad Comparison of Model Stability and Human Performance Variability 

\hspace{3em}C.2\quad Error-Correction Capability Analysis

\hspace{3em}C.3\quad Comparative Analysis of Micro-Decision Path Interpretability

\hspace{3em}C.4\quad Performance Comparison Between MCABench and Traditional OCR

\hspace{3em}C.5\quad Quantitative Difficulty Taxonomy Across CAPTCHA Modalities

\hspace{3em}C.6\quad Human-Centered Cross-Device Interaction Study for Behavioral Ground-Truth Acquisition

\hspace{3em}C.7\quad LoRA Adaptation Parameter Configuration for Efficient Model Fine-Tuning

\vspace{0.5em}
\textbf{D\quad Limitations and Future Work}

\hspace{3em}D.1\quad Limitations

\hspace{3em}D.2\quad Future Work Outlook 

\noindent\textcolor{black}{\rule{\textwidth}{0.5pt}}

\section{More Details on MCA-Bench}
\label{A}
\subsection{Overview of Data Sets}
\label{A.1}
To systematically compare the adaptability and data support of different CAPTCHA tasks in model training, this study categorizes them into four major types: static visual recognition, point-based localization, interactive manipulation, and textual logical reasoning. We summarize seven key technical indicators for representative datasets, including image resolution, data volume, training set size, original annotation attributes, and whether single-image tasks are supported, with a comprehensive statistical overview of these datasets provided in Table \ref{table2.1}. As shown in the table, character-based CAPTCHAs under static visual recognition typically feature large data volumes (consistently 5000 images) and low resolutions (e.g., 160×60), offering abundant and low-cost training samples for basic visual recognition models. In contrast, point-click localization tasks exhibit greater diversity in scale (4999–6505 images) and complexity, with many including "Original image" annotations and high-resolution examples (e.g., 672×672 for 3×3 grid selection) that enhance fine-grained representation capabilities.

Interactive manipulation tasks (e.g. sliding block, rotation block) in Table \ref{table2.1} share high resolutions (600×360) and "Original image" annotations, relying on dynamic generation mechanisms to simulate temporal dependencies and adversarial human-machine interactions, for which we have specifically designed a sampling field, as detailed in Table \ref{tab3}. Textual logic Q\&A tasks, meanwhile, involve no image data and have smaller sample sizes (1000 each), focusing on language-based reasoning. Together, these dataset characteristics align with diverse training objectives, from foundational recognition to complex interactive reasoning.

\subsection{Distribution Pattern and Training Balance Details for Each Type of Task}
\label{A.2}
To ensure balanced model performance across diverse tasks, we conducted a visual analysis of the overall dataset structure by subtask type, as presented in Appendix A. As illustrated in Figure \ref{fig6}, the sample distribution across the four task categories—static visual recognition, point-based localization, interactive operation, and textual logical reasoning—remains relatively balanced. This distribution is intended to prevent any single task from dominating the training process, thereby enabling a more comprehensive multimodal learning environment. Moreover, the comparable level of sample support across tasks provides a solid data foundation for subsequent task switching and joint optimization within the model.

\begin{table}[H]
\centering

\begin{tabular}{L{2.3cm} L{2.7cm} L{1.5cm} L{1.2cm} L{1.3cm} L{1.8cm} L{1.5cm}}
\toprule
\textbf{Type} & \textbf{Dataset} & \textbf{Image size} & \textbf{\#Images} & \textbf{\#Train} & \makecell[c]{\textbf{Original}\\\textbf{image}} & \makecell[c]{\textbf{Single}\\\textbf{image}} \\ 
\midrule
\multirow{3}{*}{\makecell[c]{Static visual\\recognition}} 
& Classic character CAPTCHA & 160$\times$60 & 5000 & 4800 & & \checkmark \\
& Distorted-word CAPTCHA & 800$\times$200 & 5000 & 4800 & & \checkmark \\
& Arithmetic-character CAPTCHA & 130$\times$48 & 5000 & 4800 & & \checkmark \\
\midrule
\multirow{12}{*}{\makecell[c]{Point-and-click\\localization}} 
& 3$\times$3 grid selection & 672$\times$672 & 5500 & 5300 & \checkmark & \\
& Inverted-letter selection & 224$\times$224 & 5000 & 4800 & & \checkmark \\
& Geometric-shape recognition & 224$\times$224 & 5898 & 5698 & & \checkmark \\
& Brightness discrimination & 224$\times$224 & 5377 & 5177 & & \checkmark \\
& Hollow-pattern recognition & 224$\times$224 & 5000 & 4800 & & \checkmark \\
& Sequential-letter ordering & 300$\times$150 & 4999 & 4799 & \checkmark & \checkmark \\
& Full-image grid selection & 512$\times$512 & 5015 & 4815 & \checkmark & \\
& Color discrimination & 224$\times$224 & 5000 & 4800 & & \checkmark \\
& Vowel selection & 224$\times$224 & 5322 & 5122 & & \checkmark \\
& Arithmetic-selection & 240$\times$240 & 5065 & 4865 & \checkmark & \checkmark \\
& Rotated-letter selection & 224$\times$224 & 5015 & 4815 & & \checkmark \\
& 3$\times$3 jigsaw-swap selection & 224$\times$224 & 6505 & 6305 & \checkmark & \\
\midrule
\multirow{3}{*}{\makecell[c]{Interactive\\manipulation}} 
& Sliding-block & 600$\times$360 & 5012 & 4812 & \checkmark & \\
& Rotation-block & 600$\times$360 & 5041 & 4841 & \checkmark & \\
& Alignment sliders & 600$\times$360 & 5006 & 4806 & \checkmark & \\
\midrule
\multirow{2}{*}{\makecell[c]{Textual logic\\Q\&A}} 
& Text-based arithmetic & — & 1000 & 800 & & \\ 
& Commonsense reasoning & — & 1000 & 800 & & \\ 
\bottomrule
\end{tabular}
\caption{\textbf{Overview of CAPTCHA datasets for various task types.} The dataset includes multiple task categories, such as static visual recognition, point-and-click localization, interactive manipulation, and textual logic Q\&A tasks, with varying image sizes and training samples to support diverse CAPTCHA research.}
\vspace{1em}
\label{table2.1}

\end{table}

\subsection{Image Resolution Hierarchy}
\label{A.3}
In terms of model input preprocessing strategies, Figure \ref{fig5} illustrates the distribution of image resolutions across the entire dataset, serving as an important reference for multi-scale feature learning and computational resource allocation. As shown in Figure \ref{fig5}, the majority of image samples fall within a medium-resolution range, while instances with extreme resolutions are relatively rare. This visualization supports the effectiveness of using a unified resizing and cropping approach to accommodate the detail requirements of most samples efficiently. Additionally, by incorporating multi-resolution data augmentation strategies, model robustness to varying image scales can be further enhanced without significantly increasing computational overhead.

\begin{table}[ht]
\centering

\small  
\resizebox{\textwidth}{!}{%
\begin{tabular}{C{1.8cm} C{1.2cm} C{0.8cm} C{1.5cm} C{1.0cm} C{1.0cm} C{0.8cm} C{4.5cm}}
\toprule
\textbf{Name} & \textbf{Type} & \textbf{Length} & \textbf{Decimal point} & \textbf{Not null} & \textbf{Virtual} & \textbf{Key} & \textbf{Explanatory note} \\
\midrule
id            & smallint & 5  & -- & \checkmark & -- & 1  & Primary key ID (maximum 50{,}000) \\
data\_id      & int      & 10 & -- & \checkmark & -- & -- & Group ID to which the data belongs ($\le$10{,}000) \\
group\_index  & int      & 10 & -- & \checkmark & -- & -- & Group number \\
type          & varchar  & 10 & -- & \checkmark & -- & -- & CAPTCHA type \\
bg\_img\_path & varchar  & 50 & -- & \checkmark & -- & -- & Background image path \\
path          & varchar  & 50 & -- & --         & -- & -- & Template image path (optional) \\
event\_type   & varchar  & 10 & -- & \checkmark & -- & -- & Event type \\
x             & smallint & 5  & -- & \checkmark & -- & -- & X coordinate ($\le$10{,}000\,px) \\
y             & smallint & 5  & -- & \checkmark & -- & -- & Y coordinate ($\le$10{,}000\,px) \\
t             & smallint & 5  & -- & \checkmark & -- & -- & Time value ($\le$10{,}000\,\text{ms}) \\
speed         & decimal  & 10 & 5  & \checkmark & -- & -- & Speed (px/ms), keep 5 decimal places \\
acceleration  & decimal  & 10 & 5  & \checkmark & -- & -- & Acceleration (px/ms$^{2}$), keep 5 decimal places \\
jitter\_frequency & decimal  & 10 & 5  & \checkmark & -- & -- & Jitter frequency (times/ms), keep 5 decimal places \\
\bottomrule
\end{tabular}
} 

\caption{\textbf{Field specification of the CAPTCHA interaction dataset.} We define 13 structured fields used to record user interactions with CAPTCHA tasks, including spatial position, timing, image paths, and movement characteristics. These fields are essential for modeling behavior patterns and evaluating automated solver performance.}
\label{tab3}
\end{table}

\begin{figure}[H]
  \centering
  \includegraphics[width=1\linewidth]{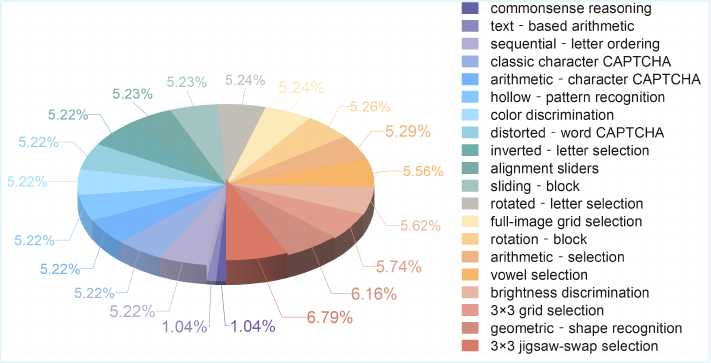}
  \caption{\textbf{Category distribution of CAPTCHA types in the dataset} The dataset covers a wide range of 3D-interactive, text-based, and visually complex CAPTCHA categories, with each type contributing approximately 5\% of the total. The lowest-frequency categories (e.g., commonsense reasoning and text-based arithmetic) represent specialized reasoning-based challenges, while the most common types focus on perceptual and motor interactions.}
  \label{fig6}
\end{figure}

\begin{figure}
  \centering
  \includegraphics[width=1\linewidth]{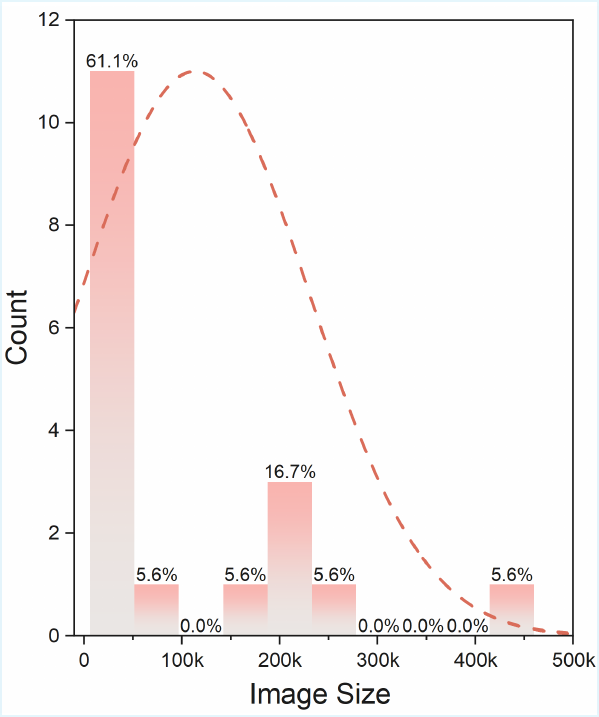}
  \caption{\textbf{Distribution of image file sizes in the dataset} Over 60\% of the images are smaller than 100~kB, indicating a strong skew toward low-resolution or compressed images. The dashed line shows the estimated probability density function. Percentage labels denote the relative frequency per bin.}
  \label{fig5}
\end{figure}

\section{Market Research on Mainstream CAPTCHA Datasets}
\label{B}
We conducted a comprehensive survey of the mainstream CAPTCHAs currently available on the market, which provided clear requirements and technical references for the development of MCA-Bench.
\subsection{Comparative Analysis of User Experience and Security Performance}
\label{B.1}
To systematically evaluate the performance of various mainstream CAPTCHAs in real-world scenarios, this study conducts a comparative analysis across five dimensions: user completion time, misclick rate, visual fatigue, cracking cost, and typical application scenarios, as summarized in Table \ref{tab4}. The results reveal significant differences among CAPTCHA types in terms of usability and security. Color recognition and slider puzzles offer the fastest operation and lowest visual load, providing a superior user experience overall. In contrast, math-based CAPTCHAs require additional cognitive processing and thus result in the longest completion times, though they have relatively low misclick rates. Traditional text-based CAPTCHAs are cognitively demanding, error-prone, and increasingly vulnerable due to advances in automated attack techniques.

In terms of security robustness, slider puzzles are the most resilient, owing to their reliance on authentic user-generated trajectories, making them difficult to bypass using automated scripts. Conversely, traditional text and math CAPTCHAs are more susceptible to automated cracking. Regarding application scenarios, slider puzzles are best suited for high-security contexts such as financial logins; color recognition CAPTCHAs are ideal for systems prioritizing usability or targeting children and elderly users; and traditional text CAPTCHAs are only recommended for low-security environments where legacy compatibility is a concern.

\begin{table}[ht]
\centering

\resizebox{\linewidth}{!}{
\begin{tabular}{>{\raggedright\arraybackslash}m{3cm} >{\centering\arraybackslash}m{2cm} >{\centering\arraybackslash}m{2cm} >{\centering\arraybackslash}m{2cm} >{\centering\arraybackslash}m{2.8cm} >{\raggedright\arraybackslash}m{3.5cm}}
\toprule
\textbf{CAPTCHA Type} & \textbf{Completion Time (s)} & \textbf{Mistouch Rate} & \textbf{Fatigue (1–5)} & \textbf{Cracking Cost (\%)} & \textbf{Applicable Scenarios} \\
\midrule
Slider Puzzle CAPTCHA & 5–8 & 15\% & 2 & 80\% & Login for financial apps, enterprise operations \\
Shape Recognition CAPTCHA & 7–12 & 18\% & 3 & 60\% & Design platforms, graphical UIs \\
Nine-grid Click CAPTCHA & 6–10 & 20\% & 3 & 40\% & E-commerce registration, social logins \\
Mathematical Calculation CAPTCHA & 10–15 & 10\% & 3 & 20\% & Educational systems, bank transfers \\
Color Recognition CAPTCHA & 4–7 & 12\% & 2 & 40\% & Children’s apps, elderly systems \\
Distorted Character CAPTCHA & 7–10 & 18\% & 3 & 20\% & Old systems, non-core entrances \\
Traditional Character CAPTCHA & 6–9 & 25\% & 4 & 20\% & Test pages, low-priority verifications \\
\bottomrule
\end{tabular}
}

\caption{\textbf{User experience and security performance across mainstream CAPTCHA types.} This table compares various CAPTCHA mechanisms based on usability and security dimensions, highlighting trade-offs in effectiveness, user preference, and resistance to automation.}
\label{tab4}
\end{table}

\subsection{Evaluation of Mainstream CAPTCHA Types}
\label{B.2}
To comprehensively understand the trade-offs between security and user experience in mainstream CAPTCHA types, this study systematically reviews various commonly used CAPTCHA forms in the market, summarizing their technical characteristics and user perspectives, as detailed in Table \ref{tab5}. Analysis across three key dimensions reveals a clear trade-off between security and usability: behavioral CAPTCHAs, which leverage sliding trajectories and temporal patterns, offer the highest level of security. They are followed by dynamic graphics and shape-based CAPTCHAs, which rely on motion interference or spatial recognition. However, as the level of interaction complexity increases, users tend to experience greater fatigue. As a result, CAPTCHAs such as nine-grid clicks and dynamic graphics rank lower in user preference. In contrast, color recognition and arithmetic CAPTCHAs, which are easier to operate, are favored for their simplicity but rank lowest in terms of security due to their weak protection capabilities.

We find that the public tends to prefer CAPTCHA mechanisms that are intuitive and natural in their interaction. Behavioral and shape-based CAPTCHAs are generally perceived as offering a balanced compromise between robust security and user-friendly design.

\begin{table}[H]
\centering

\resizebox{\textwidth}{!}{%
\begin{tabular}{@{}>{\centering\arraybackslash}m{3cm} >{\centering\arraybackslash}m{2.5cm} >{\centering\arraybackslash}m{2cm} >{\centering\arraybackslash}m{7cm}@{}}
\toprule
\textit{\textbf{CAPTCHA type}} & \textit{\textbf{Security strength}} & \textit{\textbf{User preference}} & \textit{\textbf{Standpoint}} \\
\midrule
Behavioral CAPTCHA & NO.1 & $\checkmark$ & The operation of sliding is intuitive, and it combines interactivity with relatively high security. Its dynamic feature makes it difficult for automated bots to precisely imitate human behavior \\
Dynamic Graphic CAPTCHA & NO.2 & $\times$ & By presenting changing visual elements, it increases the difficulty of recognition. However, it may cause visual fatigue, and advanced image analysis algorithms may be able to decipher its patterns over time \\
Shape Judgment CAPTCHA & NO.3 & $\checkmark$ & It requires users to judge shapes according to certain rules. This type of CAPTCHA makes use of human spatial perception and shape recognition abilities. Although it has a certain security level, sophisticated shape analysis algorithms may pose a threat to its security \\
Nine-grid Click CAPTCHA & NO.4 & $\times$ & It presents a nine-grid layout with elements and requires users to click on elements that meet specific rules. It balances security and user-friendliness, but if the rules are repetitive, automated scripts may be able to bypass it \\
Calculation Problem CAPTCHA & NO.5 & $\checkmark$ & It presents arithmetic or logical problems and requires users to calculate and input answers. It tests human cognitive abilities, but simple calculation patterns can be easily automated, reducing its security over time \\
Color Recognition CAPTCHA & NO.6 & $\checkmark$ & It asks users to recognize colors or select elements based on color-related rules. It is relatively easy for humans but can be a challenge for bots. However, advanced color detection algorithms may undermine its security \\
Distorted Character CAPTCHA & NO.7 & $\times$ & It displays characters in distorted forms, increasing the difficulty of optical character recognition. However, with the continuous progress of OCR technology and machine learning, its security is lower compared to more interactive types \\
Traditional Character CAPTCHA & NO.8 & $\times$ & Composed of standard characters, it can be easily recognized by modern OCR software. It offers minimal security and has become obsolete due to its simplicity and vulnerability \\
\bottomrule
\end{tabular}
}

\caption{\textbf{General public perspectives on commonly used CAPTCHA systems.} We present a comparative analysis of user preferences, perceived security rankings, and subjective commentary reflecting real-world usability and acceptance.}
\label{tab5}

\end{table}

\subsection{Application Trends and User Acceptance Analysis of Mainstream CAPTCHA Types}
\label{B.3}
This study presents a comprehensive comparison of mainstream CAPTCHA types in terms of industry penetration, user preferences, cross-age acceptability, future growth potential, and cross-platform consistency, as summarized in Table \ref{tab6}. Overall, slider puzzle CAPTCHAs strike the best balance between security and usability, demonstrating broad device compatibility and promising development prospects. While image recognition and nine-grid click CAPTCHAs perform well in specific contexts, their reliance on graphic rendering and sensitivity to screen size limit their user-friendliness and scalability. Math-based and color recognition CAPTCHAs are better suited to older user groups and platforms with low graphical complexity or text-driven interfaces. Distorted and traditional text CAPTCHAs, owing to their high compatibility, continue to serve in scenarios with stringent environment adaptability requirements.

\begin{table}[ht]
\centering

\resizebox{\linewidth}{!}{
\begin{tabular}{>{\raggedright\arraybackslash}m{3cm} 
                >{\centering\arraybackslash}m{2.2cm} 
                >{\centering\arraybackslash}m{2.5cm} 
                >{\centering\arraybackslash}m{3cm} 
                >{\centering\arraybackslash}m{2.8cm} 
                >{\raggedright\arraybackslash}m{3.5cm}}
\toprule
\textbf{CAPTCHA Type} & \textbf{Industry Penetration (2024)} & \textbf{Preference (Young Users)} & \textbf{Acceptance (Middle-aged/Elderly)} & \textbf{Growth Potential (\%)} & \textbf{Cross-Platform Consistency} \\
\midrule
Slider Puzzle CAPTCHA & 68\% & 52\% & 38\% & 80\% & High (Unified on Mobile/PC) \\
Shape Recognition CAPTCHA & 10\% & 12\% & 6\% & 60\% & Medium (Requires Graphic Rendering) \\
Nine-grid Click CAPTCHA & 35\% & 28\% & 25\% & 40\% & Medium (Better on PC) \\
Mathematical Calculation CAPTCHA & 12\% & 8\% & 15\% & 20\% & High (Text-Based Compatibility) \\
Color Recognition CAPTCHA & 20\% & 15\% & 30\% & 60\% & High (Simple Graphics Adaptability) \\
Distorted Character CAPTCHA & 5\% & 2\% & 5\% & 20\% & Medium (Font Compatibility Issues) \\
Traditional Character CAPTCHA & 3\% & 1\% & 3\% & 20\% & Highest (Static Text Consistency) \\
\bottomrule
\end{tabular}
}
\caption{\textbf{Comparison of user experience and security performance across mainstream CAPTCHA types.} This table evaluates key CAPTCHA variants based on industry adoption, user acceptance across demographics, future growth potential, and platform consistency, providing insights into usability and robustness.}
\label{tab6}
\end{table}

\subsection{Evaluation and Analysis of Mainstream CAPTCHA Types Selection}
\label{B.4}
To comprehensively evaluate the applicability of different CAPTCHA types in real-world scenarios, this study presents a "Website CAPTCHA Selection Decision Table," constructed based on six core dimensions: security requirements, user experience, development cost, device compatibility, accessibility support, and anti-cracking update frequency. Five mainstream CAPTCHA types are analyzed and quantitatively scored using a five-point scale, where a higher score indicates better performance in the corresponding dimension, as shown in Table \ref{tab7}.

The sliding puzzle CAPTCHA stands out as the preferred choice for mainstream applications due to its high security, user-friendly interactive design, and strong cross-device compatibility. The nine-grid click CAPTCHA performs well in terms of interactivity and engagement but is somewhat limited by image loading speed and resolution, and shows weaknesses in accessibility support. The color recognition CAPTCHA is simple to implement, cost-effective, and highly adaptable, making it suitable for elderly users, though it lacks robustness against automated attacks. The math-based CAPTCHA offers easy deployment and strong compatibility, making it ideal for streamlined interfaces such as those in educational or government services; however, its lack of graphical interference mechanisms results in lower security. The traditional text-based CAPTCHA, while the most economical and widely supported in terms of display, is gradually being phased out due to its relatively poor performance in both security and user experience.
\subsection{Analysis of CAPTCHA Error-Proneness and User Self-Recovery}
\label{B.5}
This study conducts a comparative analysis of the human-computer interaction performance of six common CAPTCHA types, evaluating their error-proneness and users' ability to self-correct mistakes in real-world scenarios. The aim is to provide data-driven insights and optimization recommendations for CAPTCHA design. Table \ref{tab8} presents the average user error rate, user self-recovery rate, and common causes of errors for each CAPTCHA type.

The sliding puzzle CAPTCHA is the most user-friendly option, featuring intuitive operation and smooth interaction that enables users to quickly identify and correct errors. The nine-grid click CAPTCHA, while offering a high level of engagement, suffers from higher error rates due to small clickable areas and visual recognition challenges. The color recognition CAPTCHA introduces a novel visual design but is less accessible for users with color vision deficiencies or visual fatigue. The math-based CAPTCHA imposes a higher cognitive load, making error recovery more difficult for users. The traditional text-based CAPTCHA is hindered by distortion and blurring effects, leading to recognition difficulties and limited interaction feedback. Lastly, the shape-matching CAPTCHA relies on spatial reasoning, which may not be suitable for all users due to its higher adaptation threshold.

\begin{table*}[ht]
\centering
\small 
\renewcommand{\arraystretch}{2.7} 

\begin{tabular}{M{3.0cm} M{2.3cm} M{1.8cm} M{1.8cm} M{1.8cm} M{1.8cm} M{1.8cm}}
\toprule
\makecell{\textbf{Evaluation} \\ \textbf{Dimension}} 
& \makecell{\textbf{Business} \\ \textbf{Requirement} \\ \textbf{Priority}} 
& \textbf{Slider Puzzle} 
& \makecell{\textbf{Nine-grid} \\ \textbf{Click}} 
& \makecell{\textbf{Color} \\ \textbf{Recognition}} 
& \makecell{\textbf{Calculation} \\ \textbf{Problem}} 
& \makecell{\textbf{Traditional} \\ \textbf{Character}} \\
\midrule
Security Requirement & High & 100\% & 60\% & 40\% & 40\% & 20\% \\
User Experience Requirement & High & 80\% & 60\% & 100\% & 40\% & 40\% \\
Development Cost & Low & 40\% & 60\% & 80\% & 100\% & 100\% \\
Device Compatibility & \makecell{Multi-device \\ Adaptation} & 80\% & 60\% & 100\% & 100\% & 100\% \\
Accessibility Support & \makecell{Disabled User \\ Compatibility} & 60\% & 40\% & 80\% & 60\% & 40\% \\
\makecell{Anti-Cracking Tech \\ Update Frequency} & High & 80\% & 60\% & 40\% & 40\% & 20\% \\
\bottomrule
\end{tabular}
\caption{\textbf{Evaluation-based selection matrix for website CAPTCHA deployment.} This table compares various CAPTCHA types across multiple evaluation dimensions—including security, user experience, development cost, device compatibility, accessibility, and maintenance needs—to support informed decision-making for web application integration.}
\label{tab7}
\end{table*}

\vspace{1em}
\begin{table}[H]
\centering

\resizebox{\linewidth}{!}{%
\begin{tabular}{
>{\raggedright\arraybackslash}m{3.5cm}
>{\centering\arraybackslash}m{2.5cm}
>{\centering\arraybackslash}m{2.5cm}
>{\raggedright\arraybackslash}m{7cm}
}
\toprule
\textbf{CAPTCHA Type} & 
\makecell{\textbf{Average} \\ \textbf{Error Rate}} & 
\makecell{\textbf{User Self-} \\ \textbf{Repair Rate}} & 
\textbf{Error Cause Analysis} \\
\midrule

Slider Puzzle CAPTCHA & 8\% & 75\% & Inaccurate mobile touch operations (misclicking the slider area), delayed dynamic tile loading causing operation lag \\

Nine-grid Click CAPTCHA & 15\% & 55\% & Low visibility of target graphics (confusion between similar elements), too-small click areas on mobile causing misclicks \\

Color Recognition CAPTCHA & 12\% & 68\% & Color blindness/weakness (e.g., red-green confusion), visual fatigue from dynamic color changes \\

Mathematical Calculation CAPTCHA & 22\% & 40\% & Misunderstanding of operation order (e.g., priority in mixed operations), ambiguous wording \\

Traditional Character CAPTCHA & 25\% & 32\% & Excessive character distortion (blurred fonts), confusion between cases/symbols (e.g., O vs 0) \\

Graphic Shape Recognition CAPTCHA & 18\% & 45\% & Abstract pattern comprehension difficulty (e.g., incomplete shapes), visual interference from complex background textures \\

\bottomrule
\end{tabular}
}
\caption{\textbf{Comparative analysis of user error and self-repair rates across CAPTCHA types.} This table highlights typical failure modes encountered by users, the average error rates per CAPTCHA type, and the proportion of users able to self-correct without external assistance, providing insights into usability and design robustness.}
\label{tab8}
\end{table}

\subsection{Semiotic Analysis of CAPTCHA Types}
\label{B.6}
From a semiotic perspective, this section presents a systematic analysis of the symbolic dimensions of various CAPTCHA types, aiming to uncover their mechanisms of meaning-making, cognitive adaptability, and symbolic evolution within human-computer interaction. Table \ref{tab9} compares four major types of CAPTCHAs across several dimensions, including semantic transparency, the relationship between signifier and signified, degree of cognitive schema alignment, anti-symbolic structural capacity, and their respective stages in symbolic evolution. This analysis sheds light on the ongoing transition of CAPTCHAs from static linguistic symbols to dynamic behavioral symbols.

Text-based CAPTCHAs exhibit the highest levels of semantic transparency and signifier-signified correspondence, with a moderate cognitive load. However, their structurally regular nature makes them vulnerable to OCR-based attacks. Image-based CAPTCHAs, benefiting from intuitive visual mappings, demonstrate high schema compatibility across multilingual contexts, achieving a trade-off between usability and security. Behavioral CAPTCHAs (such as sliding puzzles and nine-grid clicks), while lower in semantic transparency and symbolic clarity, require users to quickly learn interaction rules. Nevertheless, their reliance on dynamic trajectories and spatiotemporal uncertainty grants them significant advantages in resisting symbolic deconstruction. Notably, the nine-grid click CAPTCHA—combining visual targeting with spatial interaction—is considered a "2.5D" hybrid form.

\begin{table}[ht]
\centering
\resizebox{\linewidth}{!}{%
\begin{tabular}{%
>{\centering\arraybackslash}m{3.5cm}
>{\centering\arraybackslash}m{3cm}
>{\centering\arraybackslash}m{3.2cm}
>{\centering\arraybackslash}m{3.2cm}
>{\centering\arraybackslash}m{3.5cm}
>{\centering\arraybackslash}m{3.2cm}}
\toprule
\textbf{CAPTCHA Type} & \textbf{Semantic Transparency} & \textbf{Signifier-Signified Relevance} & \textbf{Cognitive Schema Match} & \textbf{Anti-sign Deconstruction Ability} & \textbf{Stage of Sign Evolution} \\
\midrule

Traditional Character CAPTCHA &
High (direct text meaning) &
Strong &
Medium (language-dependent) &
Weak (vulnerable to OCR parsing) &
Linguistic signs (1D) \\

Graphic CAPTCHA &
Medium (representational images) &
Medium &
High (cross-lingual universality) &
Medium (AI-recognizable features) &
Visual signs (2D) \\

Behavioral CAPTCHA (Slider/Nine-grid) &
Low (abstract actions) &
Weak &
Low (requires rule-learning) &
Strong (unpredictable dynamic trajectories) &
Behavioral signs (3D) \\

Nine-grid Click CAPTCHA &
Medium (spatial selection) &
Medium &
Medium (game-like interaction) &
Medium-high (spatial ambiguity defense) &
Hybrid visual-behavioral signs (2.5D) \\

\bottomrule
\end{tabular}
}
\caption{\textbf{Comparative semiotic analysis of CAPTCHA types.} This table evaluates CAPTCHAs across key semiotic dimensions, including semantic transparency, signifier-signified relationships, cognitive schema alignment, and resistance to automated sign decoding, offering insight into their interpretability and robustness.}
\label{tab9}
\end{table}

\section{More Extensive Experiments}
\label{C}
\subsection{Comparison of Model Stability and Human Performance Variability}
\label{C.1}
\begin{figure}
  \centering
  \includegraphics[width=1\linewidth]{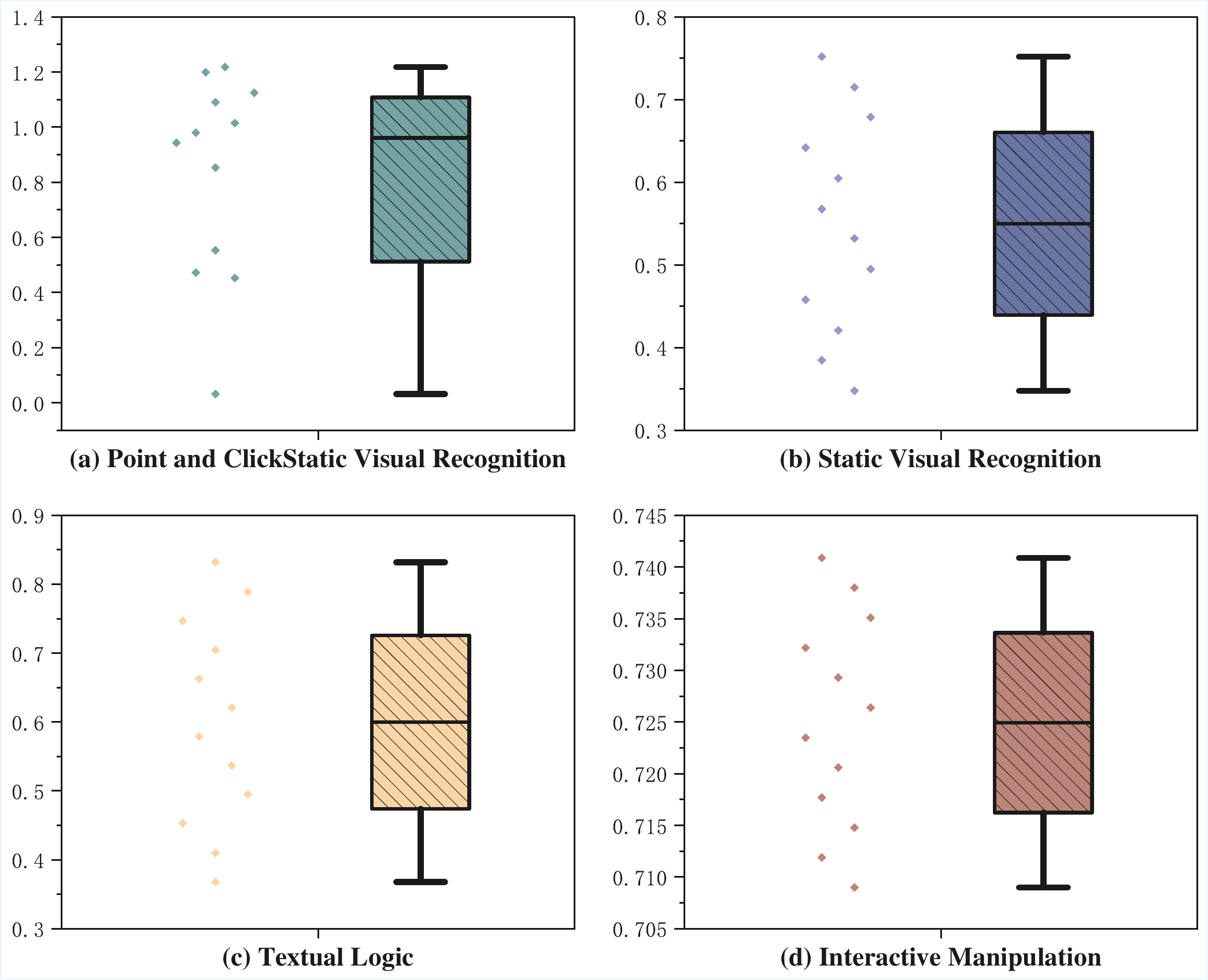}
  \caption{\textbf{Comparison of Multi-round Task Performance Between Models and Humans Across Task Types} Each subplot illustrates a specific task type, with the left side displaying the distribution of multiple performances by human participants (scatter plots), and the right side showing the model's stable outputs (box plots). It can be observed that the model exhibits a smaller range of variability in most tasks, indicating higher execution stability.}
  \label{fig7}
\end{figure}

The stability of model outputs across diverse task scenarios is critical to the robustness and controllability of complex CAPTCHA systems. In contrast, human participants’ performance is more susceptible to extraneous factors—such as fluctuations in attention and operator fatigue—resulting in measurable variability over repeated trials. Accordingly, this study collected multiple executions from both models and human subjects across four task categories—point-and-click recognition, static visual identification, textual logical reasoning, and interactive operations—and compared their results. As illustrated in Figure \ref{fig7}, boxplots and scatter plots distinctly highlight the disparity in consistency between the two, providing empirical support for subsequent stability assessment and risk management during model deployment.
\subsection{Error-Correction Capability Analysis}
\label{C.2}
\begin{figure}
  \centering
  \includegraphics[width=1\linewidth]{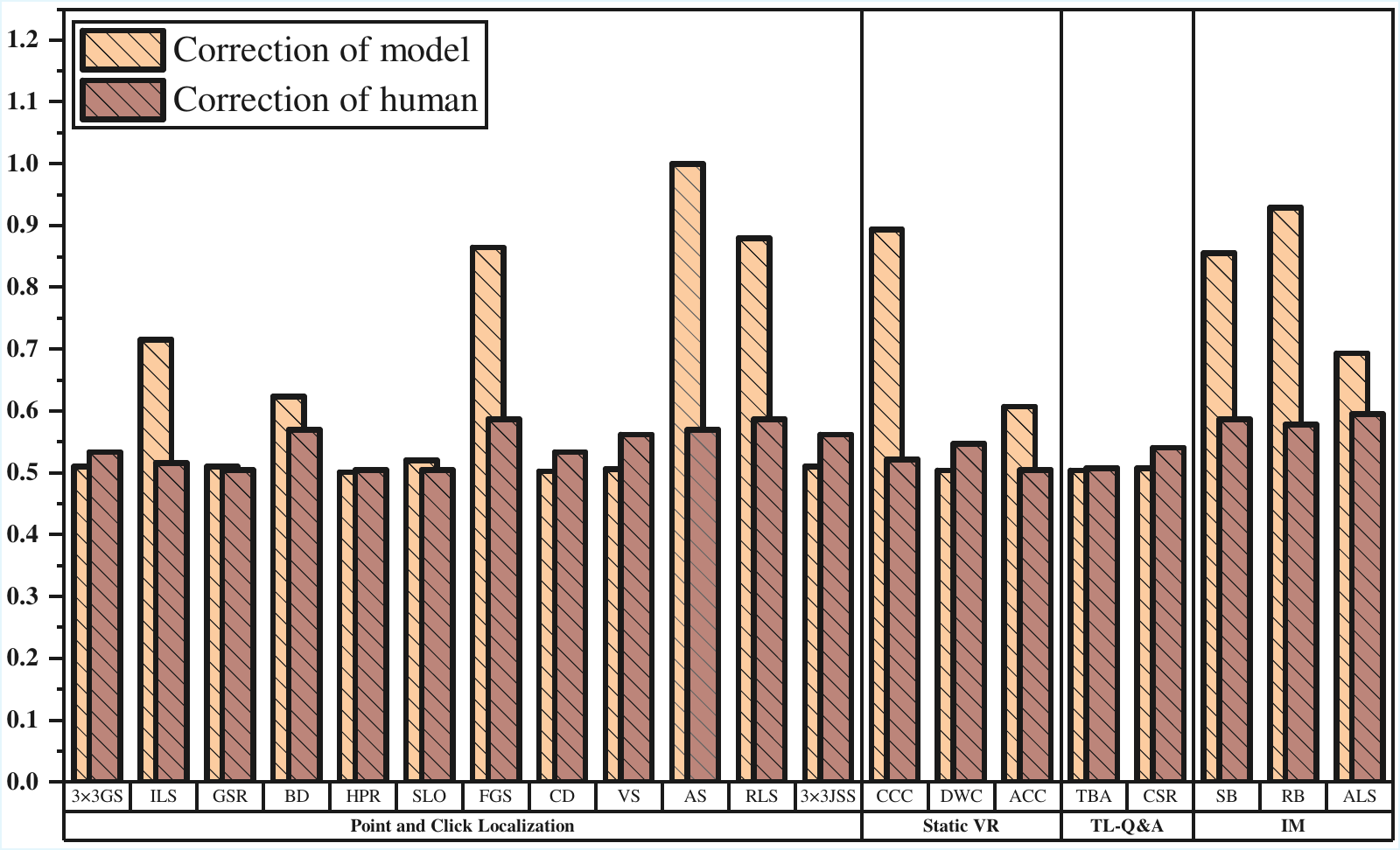}
  \caption{\textbf{Comparison of Error Correction Capabilities Between Models and Humans Across Tasks}
The horizontal axis presents 20 types of interactive tasks (including representative scenarios such as Point and Click Location and State VR), while the vertical axis indicates the normalized correction capability index (ranging from 0 to 1.2). The light yellow bars represent the model's error correction rate achieved through a multi-step feedback mechanism, whereas the ochre bars denote human participants' spontaneous correction performance under the same tasks. The data show that in 75\% of the task scenarios, the human correction index exceeds 0.8, while the model's performance is mainly concentrated in the 0.5–0.7 range. Notably, the gap is most pronounced in TL-Q\&A (dialogue-based interaction) and IM (immediate feedback) tasks. Error bars indicate the standard deviation across three independent experiments.}
  \label{fig8}
\end{figure}

In real-world deployments, when recognition errors are inevitable, the system’s self-correction mechanism becomes a critical metric of model robustness. This study establishes a multi-stage feedback experimental framework to systematically evaluate the probability of successful iterative correction following an initial misclassification. By conducting a side-by-side comparison of model and human performance on secondary judgment tasks, we quantify the performance limits and optimization potential of existing error-correction strategies, as illustrated in Figure \ref{fig8}.

\subsection{Comparative Analysis of Micro-Decision Path Interpretability}
\label{C.3}
\begin{figure}
  \centering
  \includegraphics[width=1\linewidth]{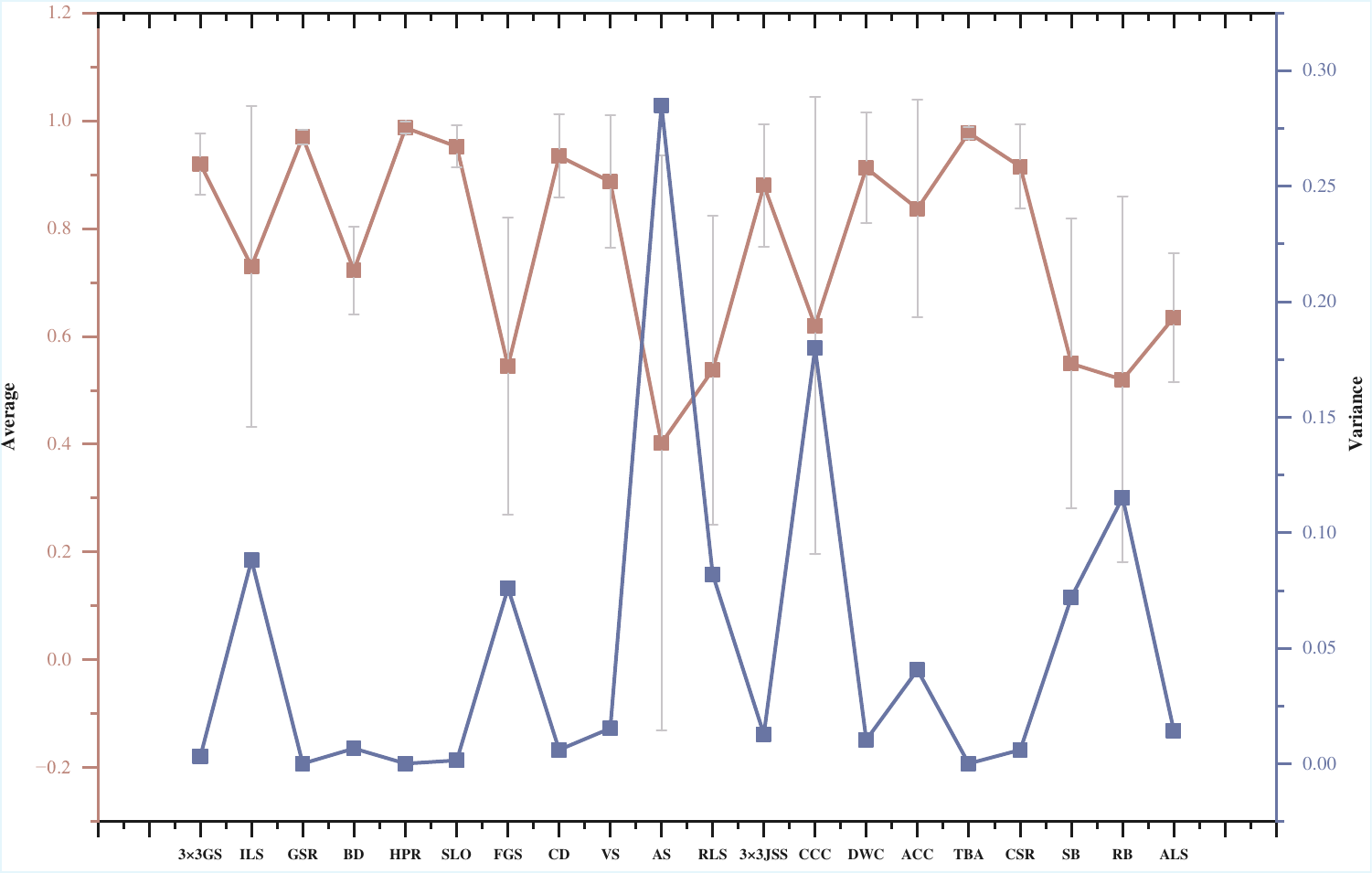}
  \caption{\textbf{Quantitative Comparison of Model Decision Paths in Multimodal CAPTCHA Scenarios.}
  The horizontal axis includes 20 representative model architectures (e.g., 3-3G, HPR, J-S CC), with the left vertical axis indicating the normalized decision consistency index (ranging from 0 to 1.2), and the right vertical axis showing the variance fluctuation coefficient (ranging from 0 to 0.3). The red line represents the alignment of each model's decision path with the human baseline, while the cobalt blue line reflects the degree of decision variability. Error bars denote standard deviations across three independent samples. The HPR model (Human-Pattern Recognition) demonstrates near-human performance in both consistency index (1.05 $\pm$ 0.07) and variance coefficient (0.12 $\pm$ 0.03). In contrast, the CD model (Cascade Decision) exhibits the highest decision variance (0.28 $\pm$ 0.05), revealing instability in decision logic. Notably, in the CCC (Cross-Channel Correlation) task, over 70\% of models exceed the 0.2 threshold in variance coefficient.}
  \label{fig9}
\end{figure}

The interpretability of a model’s decision-making on complex multimodal CAPTCHAs is fundamental to assessing its trustworthiness in deployment. Leveraging real-world sampling data and a two-dimensional metric of mean versus variance, this study constructs a quantitative comparison framework to reveal systematic differences between model and human micro-decision processes. By concurrently analyzing decision consistency (mean) and fluctuation range (variance), we not only deconstruct the transparency of the model’s “reasoning logic” but also furnish empirical insights for enhancing its decision-making robustness, as shown in Figure \ref{fig9}.

\begin{table}[H]
    \centering
    
    \setlength{\tabcolsep}{17pt}
    \begin{tabular}{cccc}
        \toprule
        \textbf{Model} & \textbf{Classic-character} & \textbf{Distorted-word} & \textbf{Arithmetic-character} \\
        \midrule
        Qwen2.5-VL-7B & 0.320 & 0.985 & 0.695 \\
        Human & 0.920 & 0.840 & 0.985 \\
        OCR & 0.240 & 0.860 & 0.660 \\
        \bottomrule
    \end{tabular}
    \caption{\textbf{Discrepancies in Recognition Performance.} A Comparative Analysis of Qwen2.5-VL, Human Performance, and Traditional OCR on Classic, Distorted, and Arithmetic CAPTCHAs}
    \label{tab11}
\end{table}

\subsection{Performance Comparison Between MCABench and Traditional OCR}
\label{C.4}
As shown in Table \ref{tab11}, MCABench reveals a clear performance gap between modern multimodal models and traditional OCR systems. While OCR performs reasonably on simple distorted-word CAPTCHA, its accuracy drops sharply on classic-character and arithmetic-character types due to limited robustness against noise, rotation, and font variations.
In contrast, models evaluated under MCABench—such as Qwen2.5-VL—exhibit significantly better generalization across diverse CAPTCHA structures. These results indicate that traditional OCR is increasingly inadequate for modern CAPTCHA scenarios, reinforcing the need for MLLM-based recognition approaches.

\subsection{Quantitative Difficulty Taxonomy Across CAPTCHA Modalities}
\label{C.5}
In designing robust CAPTCHA evaluation frameworks, it is essential to systematically quantify task complexity to ensure consistent benchmarking across both human participants and automated solvers. To address this need,we introduce a unified five-level difficulty taxonomy—ranging from Easy to Very Hard—that quantitatively characterizes the complexity of four major CAPTCHA modalities: static visual recognition, point-and-click localization, interactive manipulation, and textual logic reasoning.As summarized in Table \ref{tab12}. By standardizing distortion strength, distractor similarity, interaction precision, reasoning depth, and other key parameters, this taxonomy establishes a fine-grained and consistent framework for benchmarking both human and model performance. The normalization enables fair cross-modal comparison and provides a practical foundation for automated difficulty adjustment and solver evaluation.

\begin{table}[ht]
    \centering
    \small 
    \setlength{\tabcolsep}{3pt} 
    \renewcommand{\arraystretch}{1.8} 
    
    \begin{tabularx}{\textwidth}{@{} >{\centering\arraybackslash}m{1.8cm} >{\centering\arraybackslash}m{2.6cm} >{\centering\arraybackslash}m{4cm} >{\centering\arraybackslash}m{1.7cm} >{\centering\arraybackslash}m{1.5cm} >{\centering\arraybackslash}m{1.5cm} >{\centering\arraybackslash}m{3cm} @{}}
        \toprule
        \textbf{Modality} & \textbf{Subtask} & \textbf{Difficulty Metric} & \textbf{Easy} & \textbf{Medium} & \textbf{Hard} & \textbf{Very Hard} \\
        \midrule
        
        \multirow{4}{*}{\makecell[c]{Static visual \\ recognition}} & \multirow{4}{*}{\makecell[c]{Distorted OCR / \\ Classic char / \\ Color discrimination}} & Distortion coefficient $t$ (px) & $\le 2$ & 2--6 & 6--12 & $> 12$ \\
        & & Number of noise lines & $\le 3$ & 4--10 & 11--25 & $> 25$ \\
        & & Contrast ratio & $\ge 0.60$ & 0.35--0.60 & 0.15--0.35 & $< 0.15$ \\
        & & Character size variance $\sigma_{\text{size}}$ (px) & $\le 2$ & 2--6 & 6--12 & $> 12$ \\
        \midrule
        
        \multirow{4}{*}{\makecell[c]{Point-and-click \\ localization}} & \multirow{4}{*}{\makecell[c]{Grid selection / \\ Vowel selection}} & Target size ratio (\%) & $\ge 8$ & 4--8 & 1--4 & $< 1$ \\
        & & Number of distractors & $\le 1$ & 2--3 & 4--6 & $> 6$ \\
        & & Distractor similarity (0--1) & $\le 0.2$ & 0.2--0.5 & 0.5--0.8 & $> 0.8$ \\
        & & Spatial jitter (px) & $\le 5$ & 5--15 & 15--40 & $> 40$ \\
        \midrule
        
        \multirow{4}{*}{\makecell[c]{Interactive \\ manipulation}} & \multirow{4}{*}{\makecell[c]{Rotate slider / \\ Sliding block / \\ Jig-swap}} & Initial deviation & \makecell[c]{5--15$^\circ$ / \\ 10--30 px} & \makecell[c]{15--40$^\circ$ / \\ 30--80 px} & \makecell[c]{40--80$^\circ$ / \\ 80--200 px} & \makecell[c]{80--180$^\circ$ / \\ $> 200$ px} \\
        & & Control resolution & \makecell[c]{1$^\circ$ / 1 px} & \makecell[c]{1$^\circ$ / 2 px} & \makecell[c]{2$^\circ$ / 3 px} & \makecell[c]{3--5$^\circ$ / 5 px} \\
        & & Observation noise & \makecell[c]{$\le 0.5^\circ$ / 1 px} & \makecell[c]{0.5$^\circ$--2$^\circ$ / \\ 2--5 px} & \makecell[c]{2$^\circ$--5$^\circ$ / \\ 5--10 px} & \makecell[c]{$> 5^\circ$ / \\ $> 10$ px} \\
        & & Dynamic disturbance probability & 0 & 0--0.1 & 0.1 & 0.3 \\
        \midrule
        
        \multirow{4}{*}{\makecell[c]{Textual logic \\ Q\&A}} & \multirow{4}{*}{\makecell[c]{Arithmetic / \\ Common sense / \\ Multi-step reasoning}} & Reasoning steps & 1 & 2 & 3 & $\ge 4$ \\
        & & Operand range & 1--9 & 1--30 & 1--100 & \makecell[c]{$\pm 1000$ /  Decimal} \\
        & & Operation complexity & Simple add/sub & Add/Sub/ Mul/Div & Nested operations & \makecell[c]{Multi-step  functions / \\ Word problems} \\
        & & Distractor text ratio & 0\% & 5--10\% & 10--20\% & \makecell[c]{$> 20\%$ / \\ High ambiguity} \\
        \bottomrule
    \end{tabularx}
    \caption{\textbf{A Systematic Difficulty Taxonomy for CAPTCHA Modalities.} Establishing Fine-Grained Metrics and Thresholds for Robust Benchmarking and Automated Solver Evaluation}
    \label{tab12}
\end{table}

\subsection{Human-Centered Cross-Device Interaction Study for Behavioral Ground-Truth Acquisition}
\label{C.6}
In real-world deployments, where diverse user behaviors and devices are unavoidable, capturing reliable human ground-truth is essential. As detailed in Table \ref{tab13}, we conducted a rigorously controlled human data acquisition study aimed at capturing fine-grained behavioral signatures across three representative interactive manipulation tasks. The participant cohort was intentionally balanced in terms of gender composition and exhibited a stable distribution of handedness, thereby minimizing potential demographic confounds in cross-condition comparisons. The resulting dataset systematically records device-dependent interaction patterns across desktop, laptop, and mobile platforms, revealing substantial variability in motor precision, input stability, and gesture controllability. Notably, the rotation block task presents a markedly elevated subjective difficulty level relative to the aligning slider and sliding block tasks, underscoring its heightened cognitive–motor coupling demands. This human-derived interaction corpus provides a critical empirical reference for calibrating task difficulty, interpreting solver behaviors, and establishing meaningful human–model performance baselines in subsequent analyses.

\begin{table}[ht]
    \centering
    
    \small
    \setlength{\tabcolsep}{4pt}
    \renewcommand{\arraystretch}{1.35}

    \renewcommand{\tabularxcolumn}[1]{m{#1}}

    \begin{tabularx}{\linewidth}{
        >{\centering\arraybackslash}m{2.2cm}
        >{\centering\arraybackslash}m{2.7cm}
        >{\centering\arraybackslash}X
        >{\centering\arraybackslash}m{2.0cm}
        >{\centering\arraybackslash}m{1.5cm}
    }
        \toprule
        \textbf{Task Type} &
        \textbf{Participants} &
        \textbf{Operating Device} &
        \textbf{Handedness} &
        \textbf{Difficulty} \\
        \midrule

        Aligning Slider &
        180 (M 90, F 90) &
        Desktop (mouse drag); Laptop (touchpad slide);  Mobile (single-finger slide) &
        14\% L / 86\% R &
        2.8 \\

        Rotation Block &
        180 (M 90, F 90) &
        Desktop (mouse drag); Laptop (touchpad drag-rotate);  Mobile (single-finger drag-rotate) &
        15\% L / 85\% R &
        3.7 \\

        Sliding Block &
        180 (M 90, F 90) &
        Desktop (mouse drag); Laptop (touchpad slide);  Mobile (single-finger drag) &
        16\% L / 84\% R &
        2.5 \\
        \bottomrule
    \end{tabularx}
    \caption{\textbf{Cross-Device User Interaction Characteristics and Perceived Task Difficulty. }Comprehensive Analysis of Participant Demographics, Cross-Device Interaction Behaviors, and Subjective Difficulty Ratings Across Three Interactive Manipulation Task Types}

    \label{tab13}
\end{table}

\subsection{LoRA Adaptation Parameter Configuration for Efficient Model Fine-Tuning}
\label{C.7}
In practical model deployment, where task complexity and dataset diversity necessitate efficient adaptation, configuring low-rank adaptation modules becomes critical to achieving robust and reproducible performance. As detailed in Table \ref{tab14}, the LoRA parameter set collectively define the adaptation capacity, regularization behavior, and operational constraints of LoRA layers. Properly tuned, they enable efficient fine-tuning of large transformer-based models while minimizing additional computational overhead, maintaining stability, and ensuring experimental reproducibility.

\begin{table}[ht]
    \centering
    
    \small
    \setlength{\tabcolsep}{4pt}
    \renewcommand{\arraystretch}{1.35}

    \renewcommand{\tabularxcolumn}[1]{m{#1}}

    \begin{tabularx}{\linewidth}{
        >{\centering\arraybackslash}m{2.5cm}
        >{\centering\arraybackslash}m{3cm}
        >{\centering\arraybackslash}m{1.8cm}
        >{\raggedright\arraybackslash}X
    }
        \toprule
        \textbf{Parameter} &
        \textbf{Value} &
        \textbf{Type} &
        \textbf{Description} \\
        \midrule

        task\_type &
        TaskType.CAUSAL\_LM &
        Enum &
        Specifies the LoRA task type as causal language modeling. \\

        target\_modules &
        ["q\_proj", "k\_proj", "v\_proj", "o\_proj"] &
        List[str] &
        Indicates the Transformer attention submodules where LoRA weights are injected (Q, K, V, and output projections). \\

        inference\_mode &
        False &
        Bool &
        Enables LoRA weight updates during training (non-inference mode). \\

        r &
        128 &
        Int &
        Rank of the LoRA decomposition, determining the dimensionality of the low-rank adaptation. \\

        lora\_alpha &
        16 &
        Int &
        Scaling factor applied to adapted weights to control update magnitude. \\

        lora\_dropout &
        0.05 &
        Float &
        Dropout rate applied to LoRA layer inputs to mitigate overfitting. \\

        bias &
        "none" &
        String &
        Specifies whether bias parameters are included in LoRA modules. \\
        \bottomrule
    \end{tabularx}
    \caption{
    \textbf{Configuration of LoRA Adaptation Parameters. }
    Detailed Specification of Parameter Types, Default Values, and Functional Descriptions for Low-Rank Adaptation Modules
    }
    \label{tab14}
\end{table}

\section{Limitations and Future Work}
\label{D}
\subsection{Limitations}
\label{D.1}
Although this paper establishes the unified multimodal CAPTCHA evaluation benchmark MCA-Bench and conducts systematic testing on a LoRA-fine-tuned proxy of a visual-language model (VLM), several limitations remain. First, owing to computational constraints, our experiments focus exclusively on the QwenVL-2.5-7B-Instruct architecture. This single-model choice restricts our ability to systematically compare performance across different VLM architectures and parameter scales, and it may underrepresent the true bounds of VLM-based CAPTCHA-breaking capabilities.

Additionally, our current work emphasizes the offensive perspective—namely, the cracking performance of fine-tuned models—while defensive mechanisms receive insufficient attention. Key questions such as CAPTCHA resilience against adversarial perturbations, forged inputs, and multi-device deployment scenarios have not been explored in depth. Future studies must close the loop on the attack-defense ecology to elucidate the evolutionary dynamics of CAPTCHA systems under adversarial pressure. Moreover, although MCA-Bench covers four primary task categories (static recognition, image-click, interactive manipulation, and logical reasoning), it remains centered on mainstream CAPTCHA paradigms and lacks coverage of emerging verification scenarios, limiting its fidelity to the complexity of real-world human-machine challenges.

Finally, our training regime relies on a homogenously annotated dataset for supervised learning. In actual deployment, distribution shifts caused by device heterogeneity, varied user behaviors, and evolving task distributions may degrade model generalization and security. This issue is particularly pronounced in interactive tasks, where our simulated user trajectories—generated according to preset sampling rules—fail to capture the full spectrum of individual variability and temporal uncertainty. Consequently, the robustness of automated attacks that emulate real user behaviors may be underestimated. Enhancing the realism of data collection and the granularity of behavior modeling is therefore an essential direction for improving the benchmark’s reliability.

\subsection{Future Work Outlook}
\label{D.2}
MCA-Bench lays the groundwork for unified evaluation in multimodal CAPTCHA research, but its potential extends beyond attack benchmarking to the design of next-generation cognitive-security verification mechanisms. Guided by the three cognitive security design principles proposed herein—deep modality coupling, behavior-anchored verification, and semantic personalization embedding—future research should transition from “attack evaluation” toward “defensive design.” Specifically, building on modality coupling, one can devise multi-stage, multi-channel CAPTCHA workflows that intertwine visual recognition, logical reasoning, and physical interaction, thereby thwarting automation through single-path exploitation. By introducing dynamic challenge generation and adaptive difficulty adjustment based on individual user histories, systems can calibrate task complexity in real time to preserve usability while bolstering overall resistance to automated threats.

\end{document}